\documentclass[sn-mathphys,Numbered,iicol]{sn-jnl}% Default with double column layout

\usepackage{graphicx}%
\usepackage{multirow}%
\usepackage{amsmath,amssymb,amsfonts}%
\usepackage{amsthm}%
\usepackage{mathrsfs}%
\usepackage[title]{appendix}%
\usepackage{xcolor}%
\usepackage{textcomp}%
\usepackage{manyfoot}%
\usepackage{booktabs}%
\usepackage{algorithm}%
\usepackage{algorithmicx}%
\usepackage{algpseudocode}%
\usepackage{listings}%

\usepackage[utf8]{inputenc} % allow utf-8 input
\usepackage[T1]{fontenc}    % use 8-bit T1 fonts
\usepackage{url}            % simple URL typesetting
\usepackage{amsfonts}       % blackboard math symbols
\usepackage{nicefrac}       % compact symbols for 1/2, etc.
\usepackage{microtype}      % microtypography
\usepackage{xcolor}         % colors

\usepackage{cite}
\definecolor{citecolor}{HTML}{0071bc}
\usepackage{pdfpages}
\usepackage{makecell}
\usepackage{caption}
\usepackage{pifont}
\usepackage{capt-of}
\usepackage{subcaption}
\usepackage{array,multirow}
\usepackage[export]{adjustbox}
\usepackage{multirow}
\usepackage{soul}

\usepackage{tabu}
\usepackage{tabularx}
\usepackage{xcolor,colortbl}
\newcolumntype{L}[1]{>{\raggedright\let\newline\\\arraybackslash\hspace{0pt}}m{#1}}
\newcolumntype{C}[1]{>{\centering\let\newline\\\arraybackslash\hspace{0pt}}m{#1}}
\newcolumntype{R}[1]{>{\raggedleft\let\newline\\\arraybackslash\hspace{0pt}}m{#1}}

\def\eg{\emph{e.g.}}
\def\ie{\emph{i.e.}}

\usepackage{amsmath,amsfonts,bm}

\def\eqref#1{equation~\ref{#1}}
\def\1{\bm{1}}

\def\ve{{\bm{e}}}

\def\vw{{\bm{w}}}

\def\mA{{\bm{A}}}
\def\mB{{\bm{B}}}

\def\mE{{\bm{E}}}

\def\mG{{\bm{G}}}

\def\mK{{\bm{K}}}

\def\mQ{{\bm{Q}}}
\def\mR{{\bm{R}}}

\def\mV{{\bm{V}}}
\def\mW{{\bm{W}}}
\def\mX{{\bm{X}}}
\def\mY{{\bm{Y}}}

\DeclareMathAlphabet{\mathsfit}{\encodingdefault}{\sfdefault}{m}{sl}
\SetMathAlphabet{\mathsfit}{bold}{\encodingdefault}{\sfdefault}{bx}{n}

\definecolor{codegreen}{rgb}{0,0.5,0}
\definecolor{codeblue}{rgb}{0.25,0.5,0.5}
\definecolor{codegray}{rgb}{0.6,0.6,0.6}
\definecolor{comments}{RGB}{0,0,113}
\definecolor{red}{RGB}{160,0,0}
\definecolor{green}{RGB}{0,100,0}
\definecolor{Gray}{gray}{0.85}

\lstdefinestyle{codestyle}{
  backgroundcolor=\color{white},
  basicstyle=\fontsize{7.5pt}{7.5pt}\fontfamily{lmtt}\selectfont,
  columns=fullflexible,
  breaklines=true,
  captionpos=b,
  commentstyle=\fontsize{6pt}{7pt}\color{codegreen},
  keywordstyle=\fontsize{6pt}{7pt}\color{comments},
  stringstyle=\fontsize{6pt}{7pt}\color{red},
  showstringspaces=false,
  frame=tb,
  otherkeywords = {self},
}
\theoremstyle{thmstyleone}%
\theoremstyle{thmstyletwo}%

\theoremstyle{thmstylethree}%

\begin{document}

\title{\centering Lightweight Structure-Aware Attention \\ for Visual Understanding}


\author[1]{\fnm{Heeseung} \sur{Kwon}}\email{heeseung.kwon@inria.fr}

\author[2]{\fnm{Francisco} \sur{M. Castro}}\email{fcastro@uma.es}

\author[3]{\fnm{Manuel} \sur{J. Marin-Jimenez}}\email{mjmarin@uco.es}

\author[2]{\fnm{Nicolas} \sur{Guil}}\email{nguil@uma.es}

\author*[1]{\fnm{Karteek} \sur{Alahari}}\email{karteek.alahari@inria.fr}

\affil[1]{\orgname{Univ. Grenoble Alpes, Inria, CNRS, Grenoble INP, LJK}}

\affil[2]{\orgdiv{Department of Computer Architecture}, \orgname{University of Málaga}}

\affil[3]{\orgdiv{Department of Computing and Numerial Analysis}, \orgname{University of Córdoba}}


% \abstract{The abstract serves both as a general introduction to the topic and as a brief, non-technical summary of the main results and their implications. Authors are advised to check the author instructions for the journal they are submitting to for word limits and if structural elements like subheadings, citations, or equations are permitted.}

%%================================%%
%% Sample for structured abstract %%
%%================================%%
\abstract{
Attention operator has been widely used as a basic brick in visual understanding since it provides some flexibility through its adjustable kernels.
However, this operator suffers from inherent limitations: (1) the attention kernel is not discriminative enough, resulting in high redundancy, and (2) the complexity in computation and memory is quadratic in the sequence length. In this paper, we propose a novel attention operator, called Lightweight Structure-aware Attention (LiSA), which has a better representation power with log-linear complexity.
Our operator transforms the attention kernels to be more discriminative by learning structural patterns.
These structural patterns are encoded by exploiting a set of relative position embeddings (RPEs) as multiplicative weights, thereby improving the representation power of the attention kernels.
Additionally, the RPEs are approximated to obtain log-linear complexity.
Our experiments and analyses demonstrate that the proposed operator outperforms self-attention and other existing operators, achieving state-of-the-art results on ImageNet-1K and other downstream tasks such as video action recognition on Kinetics-400, object detection \& instance segmentation on COCO, and semantic segmentation on ADE-20K.
}




\maketitle

\section{Introduction}
% --------------------------------------------------

Since the emergence of the vision transformer (ViT)~\citep{dosovitskiy2020image}, transformers have become the dominant neural architecture for visual understanding, outperforming convolutional neural networks (CNNs).
Self-attention, a core operator of ViT, has relative merits compared to convolution because of the adjustable attention kernel and its ability to capture long-range dependencies.
However, self-attention has inherent limitations for visual recognition.
First, the attention kernel has difficulty learning discriminative features due to the lack of desirable inductive biases, resulting in high redundancy of the ViT layers~\citep{li2022uniformer,yuan2021tokens}.
Thus, it usually requires a large amount of data~\citep{dosovitskiy2020image} and aggressive augmentations~\citep{touvron2021training} to obtain good performance.
Second, the complexity of self-attention is quadratic in the length of its input sequence, making the operator impractical for high-resolution images and difficult to adopt for hierarchical models.

\begin{figure}[t]
    \centering
    \includegraphics[width=\columnwidth]{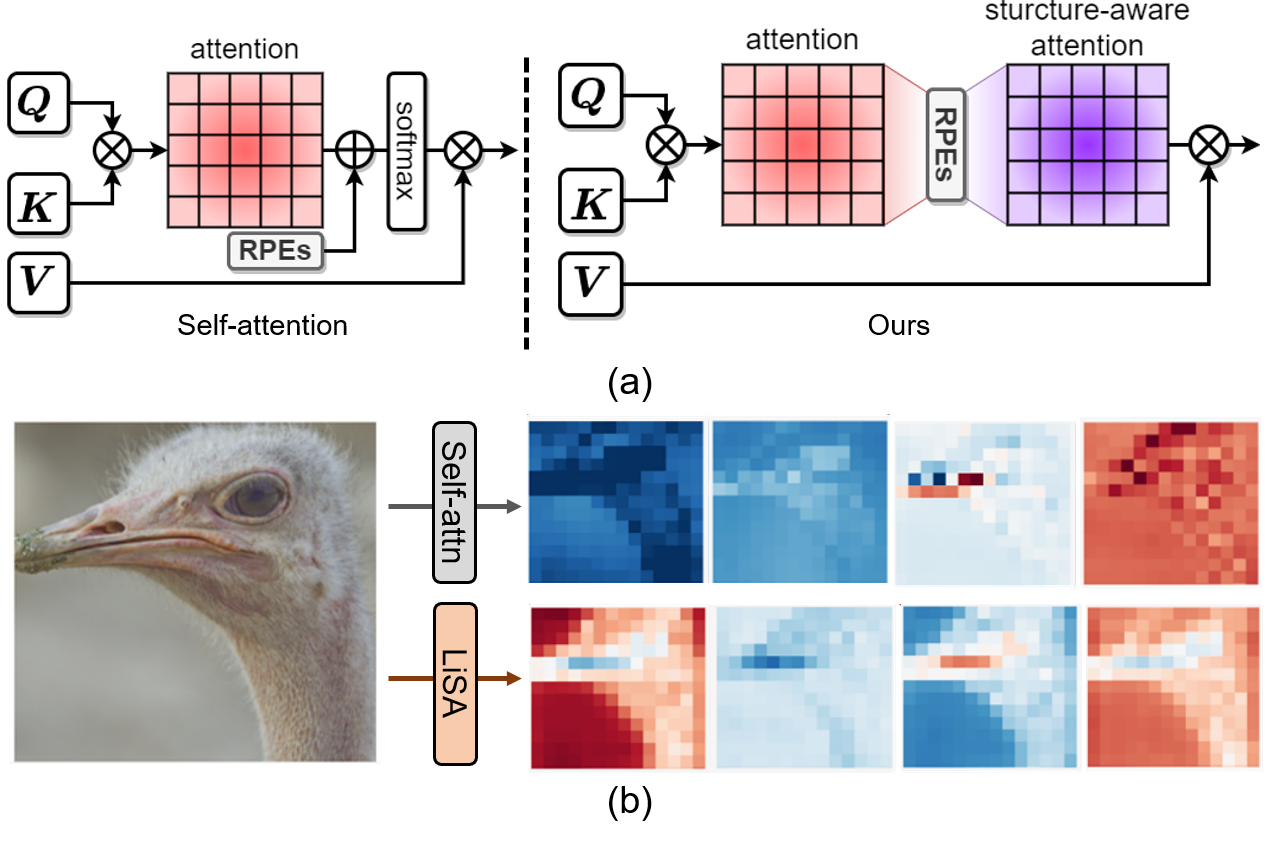}
% \beforeCaption
% \vspace{-0.8cm}
\caption{
\textbf{Self-attention vs.\ LiSA.}
(a) Process of self-attention \& LiSA: LiSA updates the attention to the structure-aware attention via RPEs.
(b) Feature visualization of self-attention \& LiSA: compared to self-attention, LiSA learns better features by capturing geometric structural patterns.
} 
\label{fig:teaser}
% \afterCaption
\end{figure}

Several approaches have proposed new types of operators to address the limitations of self-attention.
Some of them have attempted to learn better discriminative features with self-attention by including relative position embeddings (RPEs)~\citep{raffel2019exploring,liu2021swin,bello2021lambdanetworks} or capturing geometric structures (\eg,~image gradients, video motion)~\citep{zhao2020exploring,kim2021relational}.
However, these operators still have high computational complexity, which makes it challenging to capture long-range dependencies~\citep{liu2021swin,bello2021lambdanetworks,kim2021relational,zhao2020exploring}.
Some other methods have proposed efficient attention operators to handle the complexity of self-attention~\citep{wang2020linformer,choromanski2020rethinking,qin2022cosformer, liutkus2021relative,chen2021permuteformer,luo2021stable}.
Although these operators have a linear complexity with a factorized attention kernel, they often underperform, compared to the original attention~\citep{luo2021stable,qin2022cosformer}.
Recent methods have proposed %hybrid 
convolutional attention operators by integrating attention with convolution~\citep{li2022mvitv2,wu2021cvt,wang2022pvt,siinception,li2022uniformer,tu2022maxvit}.
While these operators are effective for capturing geometric structures with convolutions, their learnability is still limited since convolution kernels are local and static.
% Furthermore, they highly rely on handcrafted details (\eg, kernel shape, stride, channel ratio), which are hard to be optimized on different tasks.

In this paper, we propose an effective yet efficient operator, {\em lightweight structure-aware attention (LiSA)}.
To address the limitations of self-attention, we focus on improving the attention kernel to be more discriminative.
By leveraging the fact that the feature correlation contains rich structural information~\citep{shechtman2007matching,wang2020video,kim2021relational,kwon2021learning,sun2018pwc}, we devise a new attention operator that learns structural patterns within the query-key correlation via RPEs.
As illustrated in Fig.~\ref{fig:teaser}a, while existing methods adopt RPEs for additive interaction, we exploit RPEs as multiplicative weights.
The RPEs extract structural patterns from the attention kernel and recompose the kernel in a structure-aware manner.
% By doing this, our operator can learn both relative token orders and structural patterns.
Fig.~\ref{fig:teaser}b illustrates a few sample feature maps from the early layers of self-attention and LiSA.
LiSA effectively captures geometric structures in the image, while self-attention features are relatively weak and uninformative due to the lack of desirable inductive biases.
Finally, since the complexity of RPEs is quadratic in the sequence length, we compute them efficiently with fast Fourier transforms (FFTs), achieving log-linear complexity.

Our main contributions are as follows:
(1) we overcome the limitations of self-attention by proposing a new attention operator called LiSA, which learns structural patterns with log-linear complexity, and, 
(2) LiSANets, the models based on our LiSA operator, outperform their counterparts, achieving state-of-the-art results on visual understanding benchmarks such as ImageNet-1K~\citep{deng2009imagenet}, Kinetics-400~\citep{kay2017kinetics}, COCO~\citep{lin2014microsoft}, and ADE-20K~\citep{zhou2017ade20k}.

\section{Related Work}

\textbf{ViTs for visual understanding.}
After the success of ViT~\citep{dosovitskiy2020image}, transformer architectures have been widely adopted in a variety of visual understanding tasks~\citep{carion2020end,strudel2021segmenter,arnab2021vivit,vaswani2021scaling,yuan2021tokens,xiao2021early}. 
Several approaches have proposed improvements to the original ViT~\citep{dosovitskiy2020image}, \eg, using a teacher-student scheme~\citep{touvron2021training}, a better tokenization scheme~\citep{yuan2021tokens}, or using small splits of the tokens to obtain richer local information~\citep{han2021transformer}.
Recently, several approaches employ the hierarchical structure by adopting efficient local attention techniques~\citep{liu2021swin,zhang2021aggregating,dong2022cswin}.
However, their representation powers are still low due to the limitations of self-attention.
In this paper, we propose a new hierarchical ViT model family, LiSANets, achieving state-of-the-art performance with less computation due to the high expressivity and efficiency of LiSA.

% or downsampled attention~\citep{wang2021pyramid,fan2021multiscale,wu2021cvt,li2022mvitv2}.
% convolutions instead of self-attention operators in the early stages~\citep{dai2021coatnet,li2022uniformer},
% Others employ the hierarchical structure of CNNs to learn better discriminative features~\citep{dai2021coatnet,li2022uniformer,liu2021swin,zhang2021aggregating,wang2021pyramid,fan2021multiscale,wu2021cvt,li2022mvitv2}.
% To handle the complexity of hierarchical ViTs, these approaches use convolutions instead of self-attention operators in the early stages~\citep{dai2021coatnet,li2022uniformer}, adopt local attention~\citep{liu2021swin,zhang2021aggregating}, or downsampled attention~\citep{wang2021pyramid,fan2021multiscale,wu2021cvt,li2022mvitv2}.

\textbf{Highly-expressive operators.}
Recently proposed operators increase the representation power by developing self-attention~\citep{raffel2019exploring,liu2021swin,dai2021coatnet,zhao2020exploring,bello2021lambdanetworks,kim2021relational} or convolution~\citep{jia2016dynamic,li2021involution,chen2020dynamic,ma2020weightnet}. 
Attention-based operators have achieved this by adding relative position embeddings~\citep{raffel2019exploring,liu2021swin,dai2021coatnet} or capturing relational structures~\citep{zhao2020exploring,bello2021lambdanetworks,kim2021relational}.
Convolution-based operators have dynamically adapted convolution kernels based on the input features~\citep{jia2016dynamic,li2021involution,chen2020dynamic,ma2020weightnet}.
However, these highly-expressive operators require high computational complexity and are typically limited to local interactions~\citep{zhao2020exploring,bello2021lambdanetworks,kim2021relational,li2021involution}. One example is the relational self-attention (RSA)~\citep{kim2021relational}, which is related to our work. RSA is one of the most expressive operators that captures structural patterns with relational components, but it is also limited to local interactions due to its high computational complexity. In contrast, our proposed LiSA shows the highest level of expressivity by capturing long-range structural patterns with log-linear complexity.

\textbf{Lightweight operators.}
Some of the existing lightweight attention operators factorize the softmax attention kernel ~\citep{wang2020linformer,choromanski2020rethinking,shen2021efficient,katharopoulos2020transformers}.
While they have linear complexity, they usually perform worse than the original attention in terms of accuracy~\citep{luo2021stable,qin2022cosformer}.
Other approaches have attempted to linearize RPE-added attention operators~\citep{chen2021permuteformer,liutkus2021relative,luo2021stable}, but they still underperform on visual recognition.
Recently, a few approaches adopt FFTs for efficiently covering global receptive fields~\citep{luo2021stable,rao2021global,lee2021fnet}.
The Global Filter (GF) layer~\citep{rao2021global} is one such operator, which implements an efficient global circular convolution with FFTs.
However, the representation power of the GF layer is constrained since its static kernels hinder adaptation to general visual concepts.
Our LiSA also adopts FFTs to improve efficiency, but it focuses on learning structural patterns with its dynamic attention kernels, leading to better performance.

\textbf{Convolutional attention operators.}
Recent approaches try to combine transformers and CNNs to leverage the best of each world.
Some of them incorporate convolutions into the attention operators~\citep{wu2021cvt,d2021convit,li2022mvitv2,tu2022maxvit,li2022uniformer,siinception} to increase the expressivity of self-attention.
They often apply depthwise convolutions before computing attention to capture structural information.
Inception Mixer~\citep{siinception} splits the input channels and processes convolution and attention in parallel to increase the representation power.
While these operators are more expressive than the original attention, they highly depend on static convolution kernels for learning discriminative features.
In contrast to these, our LiSA captures discriminative features by dynamic structure-aware attention kernels, which are beneficial for learning richer visual concepts.

\section{Background}

\begin{figure*}[t]
    \centering
    \includegraphics[width=0.98\linewidth]{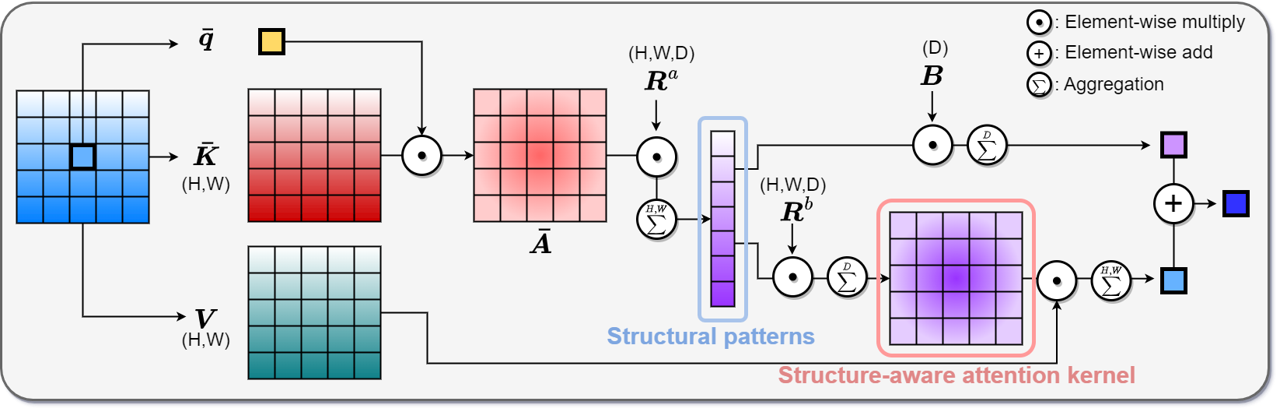}
% \beforeCaption
\caption{
\textbf{Computational graph of Structure-aware Attention (SA) for each query.} After obtaining the query-key dot-product correlation ($\bar{\mA}$), structural patterns in $\bar{\mA}$ are encoded by $\mR^a$, and utilized in two ways: 1) the patterns are used for generating a structure-aware attention kernel with $\mR^b$, and 2) directly projected as a structural feature with $\mB$. Note that $N=H\times W,C=1$ in this figure.
} \label{fig:sa}
% \afterCaption
\end{figure*}

\textbf{Self-attention.}
The self-attention operator~\citep{vaswani2017attention} is a core component in transformer architectures that generates query-key attention for updating the value.
Let $N$ denote the sequence length (the number of tokens) and $C$ the number of input channels.
Given an input feature $\mX\in\mathbb{R}^{N\times C}$, query, key, value, $\mQ,\mK,\mV\in\mathbb{R}^{N\times C}$, are first produced by independent linear projections, and each element of the output $\mY\in\mathbb{R}^{N\times C}$ of self-attention is expressed as
\begin{align}
    Y_{i,k}=\sum_{j}^N \sigma( A_{i,j}) V_{j,k},
    % \quad\textrm{where  } A_{i,j}=\frac{1}{\sqrt{C}} \sum_{k}^C Q_{i,k} K_{j,k}.
    \quad A_{i,j}=\frac{1}{\sqrt{C}} \sum_{k}^C Q_{i,k} K_{j,k}.
\end{align}
Note that $\sigma$ is the softmax function along the $j$-axis.
The two main characteristics of self-attention are that: (1) the operator represents a global interaction where the size of the attention kernel for each query is equal to $N$, and (2) the attention kernel dynamically changes according to the input feature.
However, it is unable to encode the relative order of tokens due to the lack of convolutional inductive biases~\citep{dai2021coatnet}, resulting in performance degradation on visual recognition.

\textbf{Relative position embedding (RPE).}
One of the popular schemes to handle the lack of convolutional inductive biases is adopting an RPE for the self-attention operator~\citep{raffel2019exploring,liu2021swin,dai2021coatnet}.
A common RPE has the shape of a Toeplitz matrix, and it consists of learnable weights which can be expressed as
\begin{align}
\mathcal{T}(\ve) = \left( \begin{array}{ccccc} 
e_N & e_{N+1} & e_{N+2} & \cdots   & e_{2N-1} \\
e_{N-1} & e_{N} & e_{N+1} &\cdots  & e_{2N-2} \\
\vdots & \vdots & \vdots & \ddots & \vdots \\
e_{1} & e_{2} & e_{3} &\cdots  & e_{N} 
\end{array} \right),\nonumber\\
%,\textrm{where  } \ve=\{e_{1},e_{2}, \cdots e_{2N-1}\}.
\label{eq:rpe}
\end{align}
\textrm{where  } $\ve=\{e_{1},e_{2}, \cdots e_{2N-1}\}$.
% Note that $\mathcal{T}$ shows a Toeplitz matrix of an input vector.
When RPE is added, the attention operator is formulated as
\begin{align}
Y_{i,k}=\sum_{j}^N \sigma(A_{i,j} + R_{i,j} ) V_{j,k},
% \quad\textrm{where } \mR=\mathcal{T}(\ve) \in \mathbb{R}^{N\times N}.
\ \mR=\mathcal{T}(\ve) \in \mathbb{R}^{N\times N}.
\label{eq:self_att_rpe}
\end{align}
By introducing relative positional information into attention, the self-attention operator obtains the ability to learn convolutional inductive biases.
% Indeed, the RPE provides relative positional weights to the attention kernel.

\textbf{Limitations of self-attention with RPE.}
Despite several approaches showing the effectiveness of RPE, the attention operator still has some limitations.
First, the expressivity of the operator is insufficient; it is difficult to capture geometric structures (\eg,~image gradients, video motion) since the softmax attention kernel may not be effective for encoding gradient information due to its non-negativity~\citep{kim2021relational,ramachandran2019stand}.
Second, although the attention $\mA$ suppresses photometric variations and reveals geometric structures~\citep{shechtman2007matching,kwon2021learning}, the kernel is aggregated with the value $\mV$
without leveraging structural patterns within $\mA$.
Third, the operator suffers from quadratic complexity ($\mathcal{O}(N^2)$) since the non-linear softmax function and RPE are hard to linearize.
Although a few approaches~\citep{choromanski2020rethinking,wang2020linformer,luo2021stable,qin2022cosformer} have attempted to approximate the softmax function with kernelized methods to make the operator more efficient, they do not improve over the original transformer in accuracy due to its training instability~\citep{luo2021stable} or approximation errors~\citep{qin2022cosformer}.

% !TEX root = ../main.tex
% \beforeSection
\section{Structure-aware Attention}
%\section{Our approach} \label{sec:approach}
% \afterSection

% \beforeSubsection
%\subsection{Structure-aware Attention (SA)}
\subsection{Basic Form of Structure-aware Attention (SA)}
% \afterSubsection

\textbf{Learning convolutional inductive biases.}
To handle the limitations of self-attention, we devise a new attention operator that leverages the advantages of convolution.
Unlike the conventional usage of an RPE (Eq.~\ref{eq:self_att_rpe}), we employ it as multiplicative weights for learning relative token orders as follows:
\begin{align}
    Y_{i,k}=\sum_{j}^N  \bar{A}_{i,j} R_{i,j} V_{j,k},
    % \quad \textrm{where  }   \bar{A}_{i,j}=\sum_{k}^C \bar{Q}_{i,k} \bar{K}_{j,k}.
    \quad  \bar{A}_{i,j}=\sum_{k}^C \bar{Q}_{i,k} \bar{K}_{j,k}.
    \label{eq:gdo_1}
\end{align}
Note that $\bar{\mQ},\bar{\mK}$ are L2-normalized query and key, respectively.
In Eq.~\ref{eq:gdo_1}, the RPE $\mR$ not only learns relative token orders, but also actively adjusts the attention values.
We remove the softmax function to allow the attention kernel to include negative values, which may be effective for encoding structural information.
Instead, the query and key are L2-normalized to stabilize the training procedure~\citep{luo2021stable}.
Since the matrix multiplication between the Toeplitz matrix $\mR$ and the value $\mV$ is equivalent to a global convolution~\citep{strang1986proposal} that applies the convolution kernel $\ve\in\mathbb{R}^{2N-1}$ for the value $\mV$, the operator can also be interpreted as a dynamic global convolution where the dynamic component of the convolution kernel is based on the attention $\bar{\mA}$.
Thus, the proposed operator merges the characteristics of self-attention and convolution.

\textbf{Learning structural patterns.}
Nevertheless, the above operator (Eq.~\ref{eq:gdo_1}) is still limited for capturing rich structural patterns within the attention $\bar{\mA}$.
The RPE $\mR$ is only element-wise multiplied with $\bar{\mA}$, but it cannot extract meaningful features within the query-key correlation.
To handle this, we directly extract structural patterns from $\bar{\mA}$ and regenerate the attention kernel using multiple RPEs.
The updated operator is formulated as:
\begin{align}
Y_{i,k}&=\sum_{j}^N\sum_{d}^D\sum_{n}^N (\bar{A}_{i,n} R^{a}_{i,n,d})(R^{b}_{i,j,d}V_{j,k}+B_{k,d})
    \label{eq:gdo_2}
\end{align}
where $\mR^a=\{\mathcal{T}(\ve^a_1),\mathcal{T}(\ve^a_2),\cdots,\mathcal{T}(\ve^a_D)\},\mR^b=\{\mathcal{T}(\ve^b_1),\mathcal{T}(\ve^b_2),\cdots,\mathcal{T}(\ve^b_D)\}\in\mathbb{R}^{N\times N\times D}$ are RPE tensors composed of sets of Toeplitz matrices and $\mB\in\mathbb{R}^{C\times D}$ is a learnable projection matrix, respectively.
Note that $D$ is the size of structural patterns.
The computational graph of Eq.~\ref{eq:gdo_2} is illustrated in Fig.~\ref{fig:sa}.
For each query, the learnable RPE tensor $\mR^a$ captures structural patterns by encoding the $N$-size attention kernel as a $D$-size vector.
We utilize this vector in two ways:
first, to generate a new structure-aware attention kernel along the $j$-axis using the RPE tensor $\mR^b$; second, to project it as a feature representation with the learnable matrix $\mB$.
In summary, the generated attention kernel updates $\mV$ in a structure-aware manner, and the projected feature represents the encoded structural patterns.

Note that our method differs from convolutional attention operators~\citep{li2022mvitv2,siinception,tu2022maxvit,li2022uniformer,wu2021cvt} in the way of learning discriminative features.
Convolutional attention operators process convolution and attention separately in a sequential~\citep{li2022mvitv2,li2022uniformer,tu2022maxvit,wu2021cvt} or a parallel~\citep{siinception} way.
These operators rely on convolution to obtain structural information from the input feature; however, their ability is limited because the convolution kernels are static and have restricted receptive fields.
They cannot leverage rich structural patterns within the query-key correlation.
In contrast to these operators, our structure-aware attention obtains structural information from the query-key correlation and has global receptive fields.
%Thus, structure-aware attention leverages richer structural information.

\begin{table}[t]
    \captionsetup{width=\columnwidth}
        \centering
            % \begin{tabular}[t]{L{2.2cm}|C{1.1cm}|C{1.1cm}|C{1.1cm}} 
            \scalebox{0.8}{
            \begin{tabular}[t]{lcc}             
            \toprule
            operator & computation & memory \\
            \midrule
            \midrule
            self-attention~\citep{vaswani2017attention}       
            & $\mathcal{O}(N^2C)$ & $\mathcal{O}(N^2+NC)$  \\ \midrule
            structure-aware attention (Eq.~\ref{eq:gdo_1})  & $\mathcal{O}(N^2C)$ & $\mathcal{O}(N^2+NC)$  \\
            % Eq.~\ref{eq:gdo_2}  & $\mathcal{O}(N^3+N^2C)$ & $\mathcal{O}(N^3+NC)$  \\
            structure-aware attention (Eq.~\ref{eq:gdo_2})  & $\mathcal{O}(N^2CD)$ & $\mathcal{O}(NCD)$ 
            \\
            \midrule
            \rowcolor{Gray}
            LiSA (FFT approximation)  &$\mathcal{O}(NCD\log_{2}N)$  & $\mathcal{O}(NCD)$ \\
            \midrule
            \end{tabular}
            }
% \afterTable
            \caption{\textbf{Comparison of complexity of the operators.} $N,C,D$ denote the sequence length, the number of channels, and the size of structural patterns, respectively. Our operator has log-linear and linear complexity in computation and memory, respectively.
    % We simplify the complexity terms using $M\gg L,D\leq C/L$ for ease description.
    }
\label{table_efficiency}
\end{table}

% \vspace{-0.3cm}

\begin{figure*}[t]
    \centering
    \includegraphics[width=0.98\linewidth]{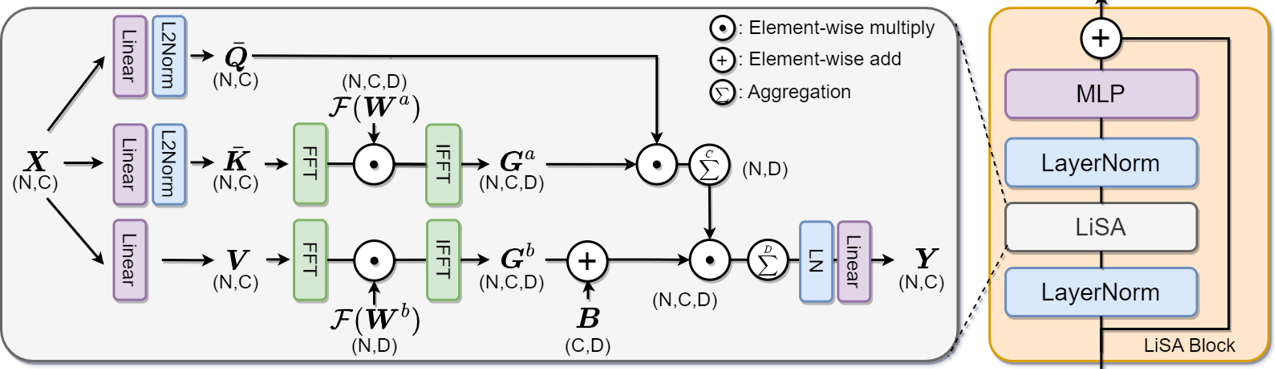}
% \beforeCaption
% \vspace{-0.2cm}
\caption{\textbf{Computational graph of LiSA and its block configuration.} See text for details.} \label{fig:lisa}
% \vspace{-0.2cm}
% \afterCaption
\end{figure*}

\subsection{Improving the Expressivity of SA} \label{sec:advanced_approach}
We can further improve its expressivity by exploiting semantic information of the input channels.
Here we describe the advanced form of our structure-aware attention.

% \textbf{Hadamard-product correlation.}
\textbf{Capturing channel-wise structural patterns.}
To exploit the semantics of the input channels, we employ a different type of query-key correlation, the Hadamard-product correlation. 
A few approaches~\citep{zhao2020exploring,kim2021relational} have demonstrated that Hadamard-product correlation is more effective than the dot-product one due to the use of richer query-key semantics.
Considering the Hadamard correlation is a 3-dimensional tensor $\bar{A}_{i,n,c}=\bar{Q}_{i,c}\bar{K}_{n,c}\in\mathbb{R}^{N\times N\times C}$, 
we expand the RPE tensor $\mR^a$ by $C$ channels for encoding the Hadamard correlation.
The modified operator is formulated as follows:
% \vspace{-4mm}
\begin{align}
    \begin{split}
    Y_{i,k} =
     \sum_{c}^C\sum_{d}^D\sum_{j,n}^N (\bar{A}_{i,n,c}\tilde{R}^{a}_{i,n,c,d})(R^{b}_{i,j,d}V_{j,k}+B_{k,d}).
     \\[-10pt] 
    \end{split}
     \label{eq:gdo_5}
% \vspace{-5mm}
\end{align}
% \vspace{-3mm}
Note that $\tilde{\mR}^a\in\mathbb{R}^{N\times N\times C\times D}$ is the expanded RPE tensor and the number of learnable weights increases from $\mE^a=\{\ve^a_1,\ve^a_2,\cdots,\ve^a_D\}\in\mathbb{R}^{(2N-1)\times D}$ to $\tilde{\mE}^a=\{\ve^a_1,\ve^a_2,\cdots,\ve^a_{CD}\}\in\mathbb{R}^{(2N-1)\times C\times D}$.
In Eq.~\ref{eq:gdo_5}, for each query, the expanded tensor $\tilde{\mR}^a$ captures channel-wise structural patterns by encoding an $N\times C$ Hadamard correlation matrix as a $D$-size vector.
Since this process does not require additional computation, the operator can exploit rich semantics through the Hadamard correlation by only increasing the number of parameters.

% \beforeSubsection
\subsection{Lightweight Structure-aware Attention (LiSA)}
% \afterSubsection
Although our operator (SA) is highly-expressive, it is not easy to apply in neural architectures due to its high computational complexity.
Here, we describe its final form, LiSA, which significantly reduces the complexity by efficiently processing the heavy RPE tensors through FFTs.

\textbf{Approximating RPEs with FFTs.}
Unlike the original attention (Eq.~\ref{eq:self_att_rpe}), the multiplicative RPEs ($\tilde{\mR}^{a},\mR^{b}$) and the absence of softmax enable the permutation of the computation order, and thus, the rearranged equation is expressed as:
% In Eq.~\ref{eq:gdo_5}, RPE tensor multiplications can be considered as follows:
% \begin{equation}
\begin{align}
    Y_{i,k}&=\sum_{c}^C \bar{Q}_{i,c} \sum_{d}^D \sum_{j,n}^N \bar{K}_{n,c}\tilde{R}^{a}_{i,n,c,d}(R^{b}_{i,j,d}V_{j,k}+B_{k,d}), \nonumber\\
    &=\sum_{c}^C \bar{Q}_{i,c} \sum_{d}^D G^{a}_{i,c,d}(G^{b}_{i,k,d}+B_{k,d}), \\
    &\textrm{where \,} \mG^a=\bar{\mK}*\tilde{\mE}^a,\mG^b=\mV*\mE^b\in\mathbb{R}^{N\times C\times D}. \nonumber
     \label{eq:gdo_6}
\end{align}
% \end{equation}
Note that $\mE^b=\{\ve^b_1,\ve^b_2,\cdots,\ve^b_D\}\in\mathbb{R}^{(2N-1)\times D}$ are learnable weights of $\mR^b$, and $*$ denotes convolution.
$\mG^b$ is a global convolution that applies the global kernels ${\mE}^b\in\mathbb{R}^{(2N-1)\times D}$ to the value $\mV$ by sharing the kernels across $C$ channels, and $\mG^a$ is a depth-wise global convolution that applies the global kernels $\tilde{\mE}^a\in\mathbb{R}^{(2N-1)\times C\times D}$ to the key $\bar{\mK}$.
In Eq.~\ref{eq:gdo_6}, improving the efficiency of SA has shifted to efficiently processing global convolutions.
To reduce the complexity of global convolutions, we approximate them as {\em global circular convolutions}~\citep{rao2021global}, indicating that the RPEs are replaced by circular position embeddings consisting of circulant matrices.
These circular convolutions can be efficiently computed by FFTs via the convolution theorem of Fourier transform: {\em multiplication in the frequency domain is equal to circular convolution in the time domain.}
Thus, we can rewrite the equation as follows:
\begin{align}
    \mG^a 
    &\approx \bar{\mK} \circledast \mW^a = \mathcal{F}^{-1}(\mathcal{F}(\bar{\mK})\odot\mathcal{F}(\mW^a)),
    \\
    \mG^b 
    &\approx \mV \circledast \mW^b = \mathcal{F}^{-1}(\mathcal{F}(\mV)\odot\mathcal{F}(\mW^b)).
    \label{eq:gdo_7}
\end{align}
Note that $\mW^a=\{\vw^a_1,\cdots,\vw^a_{CD}\}\in\mathbb{R}^{N\times C\times D}$, $\mW^b=\{\vw^b_1,\cdots,\vw^b_D\}\in\mathbb{R}^{N\times D}$ are learnable weights of the circular convolutions and $\circledast,\odot,\mathcal{F},\mathcal{F}^{-1}$ denote circular convolution, element-wise multiplication, FFT, and IFFT, respectively.
The circular convolutions have half the size of parameters since the kernel size reduces from $2N-1$ to $N$.
As shown in Tab.~\ref{table_efficiency}, we reduce the complexity in computation and memory on a log-linear scale by approximating heavy global interactions with FFTs.
The computational graph of LiSA is illustrated in Fig.~\ref{fig:lisa}.
Moreover, we further reduce the complexity by half using RFFT and inverse RFFT.
Implementing Eq.~\ref{eq:gdo_7} using standard GPU libraries has an IO bottleneck in throughput since it reads and writes the intermediate results repeatedly.
We adopt the kernel fusion strategy~\citep{fu2022hungry} to overcome this by fusing the entire computation of Eq.~\ref{eq:gdo_7} into a single kernel and computing it in SRAM.

% !TEX root = ../main.tex
% \beforeSection
\section{Experiments}
% \afterSection

We first describe the implementation details and then present extensive results.
This includes a set of comprehensive analyses and a state-of-the-art comparison on ImageNet-1K~\citep{deng2009imagenet} and ImageNet-21k~\citep{imagenet15russakovsky}. Finally, we also verify the effectiveness of LiSA on video action recognition with Kinetics-400~\citep{kay2017kinetics}, object detection with COCO~\citep{lin2014microsoft}, and semantic segmentation with ADE-20K~\citep{zhou2017ade20k}.

\subsection{Implementation details} \label{sec:implementation}

\begin{table}[t]
  \centering
 \setlength\tabcolsep{2.8pt}
\scalebox{0.85}{
% \begin{tabu}to\textwidth{lcc}\toprule
\begin{tabular}[t]{lcc}
\toprule
    % \begin{tabu}to\textwidth{l*{2}{X[c]}rr}\toprule
    Model & \#Blocks & \#Channels (\#heads) \\ 
    \midrule
    \midrule
    LiSANet-I &  12 & 192 (12) \\ \midrule
      % \midrule
    % XXNet-H-T &  [2, 2, 6, 2]  & [64 (4), 128 (8), 320 (16), 512 (32)]  & 2.5 & 16.8\\
    % LiSANet-T &  [1, 2, 5, 2] & [64 (4), 128 (8), 256 (16), 512 (32)] \\
    LiSANet-S &  [2, 4, 12, 4] & [64 (4), 128 (8), 320 (20), 384 (24)] \\
    LiSANet-B &  [4, 8, 18, 3] & [96 (6), 192 (12), 384 (24), 576 (36)] \\
     \midrule
    HyLiSANet-S &  [3, 6, 12, 4] & [64, 128, 320 (20), 384 (24)] \\
    HyLiSANet-B &  [3, 12, 18, 3] & [96, 192, 384 (24), 576 (36)] \\
      \bottomrule
    % \end{tabu}%
    \end{tabular}
    }
    \caption{Detailed configurations of different variants of LiSANet. For hierarchical models, we provide the number of channels and blocks in 4 stages.} %FLOPs are calculated with a $224\times 224$ input.}   
    \label{tab:arch}
% \vspace{-5pt}
 %\vspace{5pt}
   % \vspace{-20pt}
% \afterTable
\end{table}%

\textbf{LiSA block.} Our proposed block follows the traditional transformers sequence of layers~\citep{dosovitskiy2020image,arnab2021vivit,touvron2021training}: layer normalization (LN), attention operator, LN and MLP. Instead of using a traditional attention operator, we use LiSA.
The overall block configuration is shown in Fig.~\ref{fig:lisa}.

\textbf{LiSANet.}
To demonstrate the effectiveness of LiSA, we define three different variant architectures as shown in Tab.~\ref{tab:arch}.
The first variant is an {\em isotropic model} (LiSANet-I), which has no downsampling layers and fixes the number of tokens ($14\times 14$) at all depths.
The second is {\em hierarchical ViTs} (LiSANet-S, LiSANet-B) consisting of LiSA blocks.
All of the hierarchical ViTs are composed of 4 stages with a different number of blocks, and the number of tokens is downsampled in each stage.
And, the third is {\em hybrid ViTs} (HyLiSANet-S, HyLiSANet-B) following the strategy of recent hybrid models~\citep{li2022uniformer,dai2021coatnet,siinception}.
We adopt depthwise convolutions for the early two stages, which allows for constructing deeper layers for high-resolution stages. The structural pattern size $D$ is set to 16 for LiSANet-I and 8 for the other models.
Like many previous backbones~\citep{lin2023scale,li2023uniformer,wang2022pvt,wang2023internimage,pan2023slide,li2022mvitv2}, we adopt convolutional position embedding~\citep{chuconditional} and convolutional MLP~\citep{wang2022pvt} for our models.
All the details of our variants are summarized in Sec.A.1. of the supplementary material.

% !TEX root = ../main.tex

\begin{table*}[t]
\centering
    % \begin{minipage}{\columnwidth}
    \begin{subtable}[t]{1.1\columnwidth}
            \centering
            % \begin{tabular}[t]{L{2.2cm}|C{1.1cm}|C{1.1cm}|C{1.1cm}} 
            \scalebox{0.66}{
            % \begin{tabular}[t]{clcccc}   
            \begin{tabular}[t]{C{0.85cm}L{4.5cm}C{1.1cm}C{1.5cm}C{0.85cm}C{0.85cm}} 
            \toprule
            index & operator & FLOPs & \#params & top-1 & top-5\\
            \midrule
            \midrule
            1 & Self-attn~\citep{vaswani2017attention}           & 1.25 G & 5.72 M & 71.0 & 90.0 \\
            2 &Self-attn w/ RPE~\citep{raffel2019exploring}            & 1.25 G & 5.72 M & 72.2 & 90.9 \\
            3 &Self-attn w/ RPE ($C\uparrow$)~\citep{raffel2019exploring}            & 1.40 G & 6.44 M & 73.4 & 91.6 \\
            % \midrule
            4 &Depthwise conv ($7\times7$)~\citep{howard2017mobilenets}  & 0.84 G & 4.49 M & 69.0 & 89.2  \\
            5 &GF layer~\citep{rao2021global}          & 0.82 G & 4.90 M & 69.5 & 89.4  \\
            6 &GF layer ($C\uparrow$)~\citep{rao2021global}          & 1.27 G & 7.37 M & 72.4 & 91.0  \\
            7 &Lambda convolution~\citep{bello2021lambdanetworks}            & 2.41 G & 5.41 M & 72.6 & 91.0  \\
            8 &Convolutional attn ~\citep{li2022mvitv2}          & 1.26 G & 5.78 M & 73.3 & 91.7 \\
            9 &Convolutional attn ($C\uparrow$)~\citep{li2022mvitv2}          & 1.41 G & 6.49 M & 74.1 & 92.1            \\
            10 &Inception mixer ($C\uparrow$)~\citep{siinception}          & 1.28 G & 6.36 M & 74.4 & 92.1              \\

            11 & RSA~\citep{kim2021relational}                       & 5.34 G & 8.23 M & 74.5 & 92.2   \\   
            % SELFYNet-TSM-R50 (ours)       & 16     &   & 40.7  \\ 
            \midrule
                    \rowcolor{Gray}
            12& LiSA (ours)     &  1.21 G & 6.36 M & \textbf{74.9} & \textbf{92.4} \\     
                    % \rowcolor{Gray}
            % LiSA ($384\times384$, ours)   & 3.62 G & 7.82 M & \textbf{77.4} & \textbf{94.0} \\   
            \bottomrule
            \end{tabular}
            }
            \caption{\textbf{Comparison with other operators}.}
            \label{table_operator}
    \end{subtable}
    \begin{subtable}[t]{0.97\columnwidth}
        \centering
        \captionsetup{width=\columnwidth}
            \centering
            \scalebox{0.66}{
            % \begin{tabular}[t]{clcccc} 
            \begin{tabular}[t] {C{0.85cm}L{3.4cm}C{1.1cm}C{1.5cm}C{0.85cm}C{0.85cm}}   
            % \begin{tabular}[t]{C{1.3cm}C{1.45cm}C{1.0cm}C{1.2cm}C{0.8cm}C{0.8cm}}   
            \toprule
            index &operator & FLOPs & \#params & top-1 & top-5  \\
            \midrule
            \midrule
            1 &Self-attn  & 1.25 G & 5.72 M & 71.0 & 90.0    \\
            2 &Self-attn w/ RPE & 1.25 G & 5.72 M & 72.2 & 90.9    \\
            3 &Self-attn w/ $h$ RPEs & 1.25 G & 5.81 M & 72.7 & 91.3    \\
            \midrule
            4 &SA (Eq.~4) & 1.25 G & 5.72 M & 72.4 & 91.2    \\
            5 &SA (Eq.~4 w/ $h$ RPEs) & 1.25 G & 5.81 M & 73.6 & 91.7    \\
            % ours (Eq.~\ref{eq:gdo_5}) &  1.23 G & 6.36 M & \textbf{74.9} & \textbf{92.4}   \\
            \midrule
            6 &SA (Eq.~5) & 3.92 G & 5.99 M & 74.2 & 92.1   \\
            7& + Hadamard corr &  3.92 G & 8.09 M & 74.9 & 92.3   \\
            % \midrule
                    \rowcolor{Gray}
            8 &LiSA (ours) & 1.21 G & 6.36 M & \textbf{74.9} & \textbf{92.4}   \\
            \bottomrule
            \end{tabular}
            }
            \caption{
                        \label{table_components}
            \textbf{Effectiveness of LiSA components}.} 
            % \vspace{-3mm}
    \end{subtable}
    \begin{subtable}[t]{1.1\columnwidth}
        \centering
        \captionsetup{width=\columnwidth}
        \centering
        \scalebox{0.66}{
            % \begin{tabular}[t]{cccccc} 
        \begin{tabular}[t]{C{0.85cm}C{4.5cm}C{1.1cm}C{1.5cm}C{0.85cm}C{0.85cm}} 
            % \begin{tabular}[t]{C{2cm}C{1.0cm}C{1.2cm}C{0.8cm}C{0.8cm}}  
            % \begin{tabular}[t]{C{0.8cm}C{2.5cm}C{1.0cm}C{1.2cm}C{0.85cm}C{0.85cm}}    
        \toprule
        index & $D$ & FLOPs & \#params &  top-1 & top-5 \\
        \midrule
            \midrule
        1&~~1          &  1.11 G & 5.76 M & 71.3 & 90.5 \\
        2&~~4          &  1.13 G & 5.88 M & 73.6 & 91.6 \\
        3&~~8     &  1.17 G & 6.04 M & 74.4 & 92.2 \\
                    \midrule
        \rowcolor{Gray}
        4&~~16        &  1.21 G & 6.36 M & \textbf{74.9} & \textbf{92.4} \\
        \bottomrule
        \end{tabular}
        }
        \caption{\textbf{Effect of the structural patterns ($D$)}.} 
        \label{table_latent_dimension}
    \end{subtable}
    \begin{subtable}[t]{0.97\columnwidth}
        \centering
        \captionsetup{width=\columnwidth}
            \centering
            \scalebox{0.66}{
            % \begin{tabular}[t]{clcccc}
            \begin{tabular}[t] {C{0.85cm}C{3.4cm}C{1.1cm}C{1.5cm}C{0.85cm}C{0.85cm}}  
            % \begin{tabular}[t]{C{0.8cm}C{3.2cm}C{1.0cm}C{1.2cm}C{0.75cm}C{0.75cm}}  
            % \begin{tabular}[t]{C{3.1cm}C{1.0cm}C{1.2cm}C{0.85cm}C{0.85cm}}   
            % \begin{tabular}[t]{C{1.3cm}C{1.45cm}C{1.0cm}C{1.2cm}C{0.8cm}C{0.8cm}}             
            \toprule
            index&kernel size & FLOPs & \#params & top-1 & top-5  \\
            \midrule
            \midrule
            1&Local - $3\times3$ & 1.22 G & 5.75 M & 71.9 & 90.7    \\
            2&Local - $5\times5$ & 1.45 G & 5.80 M & 73.8 & 91.7    \\
            3&Local - $7\times7$ & 1.80 G & 5.88 M & 73.8 & 91.8    \\
                        \midrule
                    \rowcolor{Gray}
            4&Global (ours) &  1.21 G & 6.36 M & \textbf{74.9} & \textbf{92.4}   \\
            \bottomrule
            \end{tabular}
            }
            \caption{\textbf{Effect of global interactions}.} 
            \label{table_kernel_size}
    \end{subtable}
    % \end{minipage}
    % \vspace{-0.3cm}
\caption{\textbf{LiSA analysis on ImageNet-1K}. Top-1, top-5 accuracy (\%), FLOPs (G) and the number of paramaters (M) are shown.}
    \label{table_ablation}
    % \vspace{-0.1cm}
% \afterTable  
% \afterTable
\end{table*}

% !TEX root = ../main.tex

\begin{table*}[t]
    \centering
    % \vspace{-3mm}
        % \captionsetup{width=2\columnwidth}
            \centering
            % \begin{tabular}[t]{L{2.2cm}|C{1.1cm}|C{1.1cm}|C{1.1cm}} 
 \setlength\tabcolsep{2.8pt}
            \scalebox{0.73}{
            % \begin{tabular}[t]{lcccccccc} 
    \begin{tabular}[t]{lccccccccccccccc} 
            \toprule
            block &  \multicolumn{3}{c}{$(H,W)=(7,7)$} & \multicolumn{3}{c}{$(H,W)=(14,14)$} & \multicolumn{3}{c}{$(H,W)=(28,28)$} & \multicolumn{3}{c}{$(H,W)=(56,56)$}
            & \multicolumn{3}{c}{$(H,W)=(84,84)$}\\
            % \midrule
             & FLOPs  & mem & latency & FLOPs  & mem & latency & FLOPs  & mem & latency & FLOPs  & mem & latency
             & FLOPs  & mem & latency \\
             & (M)$\downarrow$ & (M)$\downarrow$ & (ms)$\downarrow$ & (M)$\downarrow$ & (M)$\downarrow$ & (ms)$\downarrow$ & (M)$\downarrow$ & (M)$\downarrow$ & (ms)$\downarrow$ & 
             (G)$\downarrow$ & (G)$\downarrow$ & (ms)$\downarrow$ & 
             (G)$\downarrow$ & (G)$\downarrow$ & (ms)$\downarrow$ \\
            % \midrule
            \midrule
            Self-attn~\citep{vaswani2017attention} 
            & 22.7 & \textbf{18.1} &\textbf{ 0.9}
            & 102.0 & 21.7 & 1.6 
            & 584.0 & 282.4 & 13.0 
            & 5.2 & 14.7 & 174.7
            & 22.2 & OOM & OOM \\
            Convolutional attn~\citep{li2022mvitv2}
            & 23.1 & \textbf{18.1} & 1.7 
            & 104.0 & 23.9 & 2.7 
            & 413.0 & 135.4 & 8.5
            & 1.6 & 2.6 & 30.5
            & 3.7 & 3.4 & 68.0 \\
            \midrule
                 \rowcolor{Gray}
            LiSA (ours)
            & \textbf{21.8} & 18.3 & 1.2 
            & \textbf{87.3} & \textbf{20.3} & \textbf{1.3} 
            & \textbf{349.0} & \textbf{87.4}& \textbf{4.6}
            & \textbf{1.4} & \textbf{0.8} & \textbf{19.8}
            & \textbf{3.1} & \textbf{1.7} & \textbf{45.0} \\  
            % \midrule
            % self-attn & 0.27 G & 889 & x & 1.28 G & 70.9 & 53.3 & 8.16 G & 67.9 & 47.2 \\
            % \midrule
            %      \rowcolor{Gray}
            % LiSA & 0.26 G & x & x & 1.02 G & x & x & 4.09 G & x & x \\  
            \midrule
            \end{tabular}
            }   
            \caption{\textbf{Comparisons among LiSA \& attention operators in FLOPs, memory, and latency.} The memory \& latency is measured by an RTX A5000 (batch=32, channels=192). OOM is an abbreviation of out-of-memory.
            % FLOPs (G), the number of parameters (M), box mAP (AP$^b$) and mask mAP (AP$^m$) are shown. Note that FLOPs are measured at resolution $800\times 1280$.
            }
            \label{table_lowres_rebuttal}
% \vspace{-5mm}
\end{table*} 

\subsection{LiSA Analysis} \label{sec:ablation_studies}
\textbf{Setup.}
We use the isotropic model (LiSANet-I) for analyses since the operators with quadratic complexity are hard to be adopted for hierarchical models due to their extreme memory consumption.
LiSANet-I is trained for 150 epochs on ImageNet1K, and we follow the rest of the training recipes suggested in~\citep{touvron2021training,liu2021swin} for a fair comparison.
Unless specified otherwise, we use $224\times224$ resolution for input.

\textbf{Comparison with other operators.}
In Tab.~\ref{table_operator}, we compare our LiSA operator with several others, including self-attention~\citep{vaswani2017attention,raffel2019exploring}, convolution~\citep{howard2017mobilenets,rao2021global}, and the other expressive operators~\citep{bello2021lambdanetworks,kim2021relational,li2022mvitv2,siinception}.
For a fair comparison, we only replace our operator with those others in the LiSA blocks, and all the receptive fields are set as global, except for the $7\times 7$ depthwise convolution~\citep{howard2017mobilenets}.
LiSA substantially outperforms self-attention with and without RPE~\citep{raffel2019exploring} (first two rows) in accuracy, showing the impact of learning structural patterns.
The accuracy of LiSA is even 1.5\% higher than the self-attention with larger channels (3$^{\textrm{rd}}$ row), indicating that the gain does not come from the increased parameters.
LiSA also performs better than the GF layer~\citep{rao2021global} (5, 6$^{\textrm{th}}$ rows), which is equal to a global circular convolution.
This implies that the adjustable attention kernel of LiSA is a better fit for learning various visual concepts rather than the static convolution kernel.

For the convolutional attention operator (8, 9$^{\textrm{th}}$ rows), we adopt multi-scale attention~\citep{li2022mvitv2}, one of the advanced operators.
The operator applies a $3\times 3$ depthwise convolutions for the input features before computing attention, and the stride is set to $1\times 1$.
Inception mixer~\citep{siinception} splits input channels and processes convolution and attention in parallel for learning both high and low-frequency features, and the stride is set to $1\times 1$.
To demonstrate fair comparisons between these advanced attention operators with ours, we match the number of parameters and FLOPs in a similar scale by increasing the number of channels (9, 10$^{\textrm{th}}$ rows).
Convolutional attention and Inception mixer perform better than self-attention in accuracy, but their accuracies are lower than that of LiSA, even with larger FLOPs and number of parameters.
We conjecture that LiSA learns better discriminative features by capturing structural patterns inside the attention.
RSA~\citep{kim2021relational} shows high accuracy by learning structural patterns via its relational components, but it requires a significantly large computation budget due to the larger correlation.
In contrast, LiSA shows the best trade-off between accuracy and FLOPs, achieving the best accuracy among the operators with lower FLOPs.

\textbf{Effectiveness of LiSA components.}
In Tab.~\ref{table_components}, we provide a detailed analyses of the components of LiSA.
We first compare how convolutional inductive biases are learned in self-attention by our (Eq.~4) parameters.
For the same FLOPs and the number of parameters, our approach that uses an RPE as multiplicative weights (``SA (Eq.~4)'', $4^{\textrm{th}}$ row in the table) is better than standard self-attention with RPE ($2^{\textrm{nd}}$ row) in accuracy.
The accuracy gap between self-attention and ours becomes more clear when we use independent learnable weights (RPE) for each attention head ($3^{\textrm{rd}}$ vs.\ $5^{\textrm{th}}$ row), thus showing that our RPEs defined in Eq.~4 are more beneficial than those of standard self-attention.
This indicates that our attention containing negative values is potentially more effective for learning spatial features such as gradient information compared to softmax attention.
Next, we demonstrate our structure-aware attention variants in the third part of the table.
Comparing Eq.~4 with Eq.~5 ($4^{\textrm{th}}$ vs.\ $6^{\textrm{th}}$ row), we validate the effectiveness of learning structural patterns, which improves by $1.8\%$ in top-1 accuracy.
With the Hadamard correlation, our operator improves the top-1 accuracy by $1.0\%$ without any additional FLOPs.
Lastly, LiSA with FFT approximation dramatically reduces the computational cost ($3\times$) without compromising accuracy.

\textbf{Effect of the structural patterns $\textbf{\textit{D}}$.}
Our $D$ parameter represents the size of the encoded vector that learns structural patterns from the query-key correlation. Thus, larger values are expected to result in better performances. Tab.~\ref{table_latent_dimension} shows the impact of the structural patterns by varying the values of $D$.
As shown in Tab.~\ref{table_latent_dimension}, the size of $D$ significantly impacts the accuracy, which implies the importance of structural patterns.
The accuracy is indeed improved with larger values of $D$, $16$ being the best one.
Compared to the results of Tab.~\ref{table_operator}, the case of $D=8$ already surpasses all the other operators in accuracy except for RSA.
Note that the cases over $D=16$ are not reported since the accuracy becomes saturated.

\textbf{Effect of global interactions.}
Tab.~\ref{table_kernel_size} studies the influence of local and global LiSA kernels.
Focusing on the local LiSA kernels (first three rows in the table), the larger the kernel, the larger the number of parameters, FLOPs, and accuracies. This shows that a big kernel with more trainable parameters produces better results but at an increased computational cost. However, our global version (last row) allows a larger number of parameters with smaller FLOPs due to FFT, resulting in the best performance among all the variants.

% !TEX root = ../main.tex

\begin{table}[t]
  \centering
           \scalebox{0.75}{
    % \begin{tabular}[t]{clccc}
    \begin{tabular}[t]{C{1.0cm}L{3.9cm}C{1.0cm}C{1.2cm}C{0.8cm}} 
    % \begin{tabular}{L{2.2cm}|C{1.1cm}|C{1.1cm}|C{1.1cm}} 
    \toprule
    type & model & FLOPs & \#params  & top-1  \\ \midrule
    \midrule
    % ResNet-18~\citep{he2016deep} & 12    & 1.8    &  69.8 & 89.1  \\
    % % %  RegNetY-1.6GF~\citep{regnet} & 11    & 1.6      & 78.0  & - \\
    % PVT-Ti~\citep{liu2021pay} & 13     & 1.9     & 75.1  & - \\
    % % %  Deit-Ti~\citep{touvron2021training} & --    & ---   &  -- & ---  \\
    % GFNet-H-Ti~\citep{rao2021global} & 15 & 2.1 & 80.1 & 95.1 \\\midrule
    % Ours-H-Ti & 17 & 2.5 & xx.x & xx.x \\\midrule
     & ResNet-50~\citep{he2016deep}&  4.1 G& 26 M &  76.1   \\
     & RegNetY-4.0GF~\citep{radosavovic2020designing} & 4.0 G &  21 M    & 80.0  \\
     CNN & GFNet-H-S~\citep{rao2021global} &  4.6 G & 32 M &  81.5 \\ 
     & ConvNext-T~\citep{liu2022convnet} &  4.5 G & 29 M &  82.1 \\ 
     & InternImage-T~\citep{wang2023internimage} & 5.0 G & 30 M & 83.5 \\
     \midrule
     & PVT-S~\citep{wang2021pyramid}&   3.8 G & 25 M & 79.8  \\
     & Deit-S~\citep{touvron2021training}&  4.6 G & 22 M &  79.9   \\
    
    ViT & Swin-Ti~\citep{liu2021swin}&  4.5 G & 29 M & 81.2  \\
     & T2T-ViT-14~\citep{yuan2021tokens}&  4.8 G & 22 M &  81.5  \\
     & CSwin-T~\citep{dong2022cswin}&  4.3 G & 23 M & 82.7  \\  
              \rowcolor{Gray}
     % & LiSANet-S (ours)  & 3.5 G & 23 M & \textbf{82.9}  \\
     & LiSANet-S (ours)  & 3.9 G & 25 M & \textbf{83.4}  \\
            \midrule
     & CvT-13~\citep{wu2021cvt} &  4.5 G & 20 M &  81.6   \\
     & CoAtNet-0~\citep{dai2021coatnet}& \ 4.2 G & 25 M &  81.6  \\
     & MViTv2-T~\citep{li2022mvitv2}&  4.7 G & 24 M &  82.3  \\
     Hybrid & iFormer-S~\citep{siinception}&  4.8 G & 20 M & 83.4  \\
      & Slide-PVTv2-B2~\citep{pan2023slide} & 4.2 G & 23 M & 82.7 \\
     & UniFormer-S~\citep{li2023uniformer} & 3.6 G & 22 M & 82.9 \\

     % \midrule

     % LiSANet-H-S (ours)  & & 2.9 G & 19 M & \textbf{82.5}  \\
     % LiSANet-H-S (ours)  & & 2.9 G & 19 M & \textbf{82.5}  \\
     % & HyLiSANet-S (ours)   & 4.5 G & 23 M & \textbf{83.4}  \\
     & SMT-S~\citep{lin2023scale} & 4.7 G & 21 M & \textbf{83.7} \\          \rowcolor{Gray}
     & HyLiSANet-S (ours)   & 4.5 G & 24 M & \textbf{83.7}  \\
     \midrule
     \midrule
    & ResNet-101~\citep{he2016deep} &   7.9 G & 45 M &   77.4   \\
      &RegNetY-8.0GF~\citep{radosavovic2020designing}  &  8.0 G & 39 M & 81.7   \\
     CNN &GFNet-H-B~\citep{rao2021global} &  8.6 G & 54 M & 82.9  \\
    &ConvNext-B~\citep{liu2022convnet} &   15.4 G & 89 M &  83.8   \\
     & InternImage-S~\citep{wang2023internimage} & 8.0 G & 50 M & 84.2 \\
     \midrule
     &PVT-L~\citep{wang2021pyramid} &   9.8 G & 61 M     & 81.7   \\
     &Deit-B~\citep{touvron2021training} & 17.5 G & 86 M     &  81.8   \\ ViT
     &T2T-ViT-24~\citep{yuan2021tokens}&  13.8 G & 64 M &  82.3  \\
     &Swin-B~\citep{liu2021swin} &   15.4 G & 88 M & 83.5 \\
     &CSwin-B~\citep{dong2022cswin}&  15.0 G & 78 M & 84.2  \\
     \rowcolor{Gray}
     &LiSANet-B (ours) &  10.4 G & 51 M & \textbf{84.6} \\  
                \midrule
     & UniFormer-B~\citep{li2023uniformer} & 8.3 G & 50 M & 83.9 \\
     &CoAtNet-2~\citep{dai2021coatnet}&  15.7 G & 75 M &  84.1   \\
     &VAN-B4~\citep{guo2022visual}&  12.2 G & 60 M & 84.2  \\
    % &BiFormer-B~\citep{zhu2023biformer}&  9.8 G & 57 M & 84.3  \\
     & Slide-Swin-B~\citep{pan2023slide} & 15.5 G & 89 M & 84.2 \\
      &MViTv2-B~\citep{li2022mvitv2}&  10.2 G & 52 M & 84.4  \\
     Hybrid &MaxViT-S~\citep{tu2022maxvit}&  11.7 G & 69 M & 84.5 \\
     & iFormer-B~\citep{siinception} & 9.4 G & 48 M & 84.6  \\
     & SMT-L~\citep{lin2023scale} & 17.7 G & 81 M & 84.6 \\
      & Slide-CSwin-B~\citep{pan2023slide} & 15.0 G & 78 M & 84.7 \\

     &iFormer-L~\citep{siinception}&  14.0 G & 87 M & 84.8  \\

     % \midrule
    %  \bottomrule
 \rowcolor{Gray}
     % &HyLiSANet-B (ours)  & 12.9 G & 57 M & 84.8  \\  \rowcolor{Gray}
     % &HyLiSANet-B (ours, 256$\times$256)  & 16.7 G & 57 M & \textbf{85.0}  \\
     &HyLiSANet-B (ours)  & 11.7 G & 50 M & \textbf{85.0}  \\
     \midrule
    \end{tabular}%
    }
    \caption{\textbf{Comparison to the state-of-the-art models on ImageNet-1K.}
% The `attn \& conv' column indicates if models are using both self-attention and convolution.
FLOPs (G), the number of parameters (M), top-1 accuracy (\%) on the ImageNet validation set are shown. All the models use $224\times 224$ resolution images.} %\vspace{5pt}
  \label{tab:main2}
% \vspace{-0.3cm}
%   \vspace{-10pt}
% \afterTable
\end{table}%

\textbf{Efficiency of the LiSA block.}
In Tab.~\ref{table_lowres_rebuttal}, we demonstrate the efficiency of LiSA in terms of FLOPs, memory consumption, and latency.
We compare our LiSA block with a standard attention block and measure the performance of a single block ($C=192$) by varying the number of tokens.
We also compare the LiSA block with a convolutional attention block~\citep{li2022mvitv2}, which downsamples the query and key to $14\times14$ before computing attention by depthwise convolutions with multiple strides.
This attention block is more efficient than the standard one on high-resolutions (\eg,~$28\times28$, $56\times56$), but it becomes less efficient on low-resolutions due to added depthwise convolutions.
% This attention block is more efficient than the standard one in most cases, but its FLOPs, latency, and memory usage increase exponentially since its computational complexity is still quadratic in the number of tokens.
In comparison, the efficiency values of LiSA increase gracefully due to the log-linear complexity, and thus LiSA achieves the best computational performance except for $7\times7$, where LiSA is slightly slower due to a higher number of sequential operations.
Additional details are presented in the supplementary material.

\subsection{Image classification}
\textbf{ImageNet-1K.}
We train our hierarchical models (LiSANet-S, LiSANet-B, HyLiSANet-S, HyLiSANet-B) for 300 epochs, and follow the rest of the training recipes suggested in~\citep{touvron2021training,liu2021swin} for a fair comparison.
In Tab.~\ref{tab:main2}, we compare our hierarchical models with state-of-the-art approaches on ImageNet-1K, including CNNs~\citep{he2016deep,radosavovic2020designing,rao2021global}, ViTs~\citep{touvron2021training,yuan2021tokens,wang2021pyramid,liu2021swin,dong2022cswin}, and hybrid models containing both convolution and self-attention~\citep{wu2021cvt,dai2021coatnet,li2022mvitv2,siinception,tu2022maxvit,pan2023slide,li2023uniformer,wang2023internimage,lin2023scale}.
In the top half of the table, we present the results of small models with comparable FLOPs and number of parameters.
Our pure ViT model, LiSANet-S, clearly outperforms all the other ViTs in terms of accuracy and FLOPs, demonstrating the effectiveness of LiSA.
Compared with other hybrid models, our HyLiSANet-S achieves better or similar results than other approaches with comparable FLOPs and number of parameters.
% Models that incorporate convolutions into ViTs~\citep{wu2021cvt,dai2021coatnet}
% perform better than the others on the accuracy measure.
% Our proposed model, LiSANet-S, clearly outperforms all the other models in terms of both accuracy and FLOPs.

In the case of larger models, grouped in the bottom half of the table, again our pure ViT model, LiSANet-B, obtains the best accuracy with the lowest FLOPs and number of parameters, showing the benefits of our LiSA operator. Focusing on the hybrid models, our proposed HyLiSANet achieves the best accuracy ($85.0\%$) with much lower FLOPS and number of parameters than other approaches.
Thus, our proposed models achieve state-of-the-art results, especially for large models, improving traditional ViTs and more complex models that rely on additional techniques such as window attention (Swin~\citep{liu2021swin}), depthwise convolutions plus self-attention (MViTv2~\citep{li2022mvitv2}, SMT~\citep{lin2023scale}) or processing convolution and attention in parallel (iFormer~\citep{siinception}). This indicates an essential difference between convolution and LiSA. Convolution-based models converge fast and are data-efficient by their strong inductive biases such as 2D locality, but their performance is restricted when the models are scaled up since convolution may hinder adaptation to general visual concepts due to its static kernels. Whereas, our model achieves great performance even with a larger scale since LiSA can adaptively aggregate spatial information by its structure-aware attention kernels.

\begin{table}[t]
    \centering
        \captionsetup{width=0.98\columnwidth}
            \centering
            % \begin{tabular}[t]{L{2.2cm}|C{1.1cm}|C{1.1cm}|C{1.1cm}} 
            \scalebox{0.76}{
            % \begin{tabular}[t]{lcccc} 
    \begin{tabular}[t]{L{3.0cm}C{1.5cm}C{1cm}C{1.2cm}C{0.8cm}} 
            \toprule
            model &  img size & FLOPs & \#params & top-1  \\
            \midrule
            \midrule
            ConvNext-T~\citep{liu2022convnet}     & 224$\times$224 & 4.5 G & 29 M & 82.9 \\
            ViT-B/16~\citep{dosovitskiy2020image}     & 384$\times$384 & 55.4 G & 88 M & 84.0 \\  
            ConvNext-S~\citep{liu2022convnet}     & 224$\times$224 & 8.7 G & 50 M & 84.6 \\
            Swin-B~\citep{liu2021swin}     & 224$\times$224 & 15.4 G & 88 M & 85.2 \\
            ConvNext-B~\citep{liu2022convnet}     & 224$\times$224 & 15.4 G & 89 M & 85.8 \\
            CSwin-B~\citep{dong2022cswin}     & 224$\times$224 & 15.0 G & 78 M & 85.9 \\          
            % ConvNext-T~\citep{liu2022convnet}     & 384$\times$384 & 13.1 G & 29 M & 84.1 \\
            % ConvNext-S~\citep{liu2022convnet}     & 384$\times$384 & 25.5 G & 50 M & 85.8 \\
            % Swin-B~\citep{liu2021swin}     & 384$\times$384 & 47.1 G & 88 M & 86.4 \\
            % ConvNext-B~\citep{liu2022convnet}     & 384$\times$384 & 45.1 G & 89 M & 86.8 \\
            % CSwin-B~\citep{dong2022cswin}     & 384$\times$384 & 47.0 G & 78 M & 87.0 \\
                        \midrule
                 \rowcolor{Gray}
            HyLiSANet-S     & 224$\times$224 & 4.5 G & 24 M & 84.3 \\                 \rowcolor{Gray}
            HyLiSANet-B    & 224$\times$224 & 11.7 G & 50 M & \textbf{86.2} \\
            % HyLiSANet-S     & 384$\times$384 & x G & 23 M & x \\
            % HyLiSANet-B    & 384$\times$384 & x G & 57 M & x \\
            \midrule
            \end{tabular}
            }
            % \vspace{-0.3cm}
            \caption{\textbf{Comparison with other models pre-trained on ImageNet-21K and fine-tuned on ImageNet-1K}. Top-1 accuracy (\%), FLOPs (G), and the number of parameters (M) are shown.}
            \label{tab:21k}
     % \afterTable
    % \vspace{-0.4cm}
\end{table}

\textbf{ImageNet-21K.}
We have demonstrated the results with HyLiSNets fine-tuned from ImageNet-21k pre-training in Table~\ref{tab:21k} to check the behavior of our models in the large-scale data regime.
We train 90 epochs for ImageNet-21K pre-training, and fine-tune 30 epochs on ImageNet-1K. We follow the setup of ConvNext~\citep{liu2022convnet} models for a fair comparison, and all the other details are summarized in Sec.A.2 of the supplementary material.
When our models are pre-trained on ImageNet-21K, the accuracies are improved substantially. HyLiSANet-S \& -B obtain 0.9\% \& 1.4\% accuracy gains compared to the previous ImageNet-1K results, respectively.
Considering iFormer~\citep{siinception} could not be trained on ImageNet-21K due to the manually defined channel ratio~\citep{siinception}, these results demonstrate our models can learn general visual concepts when they can access a larger amount of training data.
Both HyLiSANet-S and HyLiSANet-B outperform the other models with lower computation, showing the superior scalability of HyLiSANets.

\begin{table}[t]
\centering
% \vspace{-3mm}
\small
 \setlength\tabcolsep{2.8pt}
% \captionsetup{width=0.88\textwidth}
\renewcommand{\arraystretch}{0.9}
            \scalebox{0.73}{
\begin{tabular}{lcccccc}
\toprule
model   &  img size &FLOPs   & \#params & mem & imgs/s & top-1 \\ 
\midrule
% MobileOne-S4 (CVPR'23) & 224$^2$ & 3.0 G  &  15 M &  1.2 G  & 1250.0 & 79.4 \\
Swin-B~\citep{liu2021swin} & 224$\times$224 & 15.4 G  &  88 M &  1.8 G  & 306.7 & 83.5 \\
Swin-B~\citep{liu2021swin} & 384$\times$384 & 47.1 G  &  88 M &  2.9 G  & 96.5 & 84.5 \\
ConvNext-B~\citep{liu2022convnet} & 224$\times$224 & 15.4 G  &  89 M &  1.8 G  & 312.5 & 83.8 \\
ConvNext-L~\citep{liu2022convnet} & 224$\times$224 & 34.4 G  &  198 M &  3.1 G  & 171.2 & 84.3 \\
FastViT-SA36~\citep{vasu2023fastvit}& 384$\times$384 & 12.6 G  &  30 M &  4.2 G  & 194.6 & 84.5 \\
FastViT-MA36~\citep{vasu2023fastvit}& 384$\times$384 & 17.7 G  &  43 M &  5.0 G  & 168.4 & 84.9 \\
 \midrule
%   \rowcolor{Gray}
% LiSANet-S & 224$^2$ &3.5 G & 23 M & 1.4 G & 806.5 & 82.9 \\ 
 \rowcolor{Gray}
LiSANet-B & 224$\times$224 &10.4 G & 51 M & 2.0 G & 347.2 & 84.4 \\ 
 \rowcolor{Gray}
HyLiSANet-B & 224$\times$224 & 11.7 G & 50 M &  1.8 G & 299.4 & 85.0 \\ 

\bottomrule
\end{tabular}
}
% \vspace{-3mm}
% \captionsetup{font=small} 
\caption{\textbf{Throughput comparison among modern ViT models on ImageNet-1K.} For the images per second (imgs/s) metric, higher is better.} \label{table_throughput}
% \caption{Throughput comparison on an RTX A6000 with batch size of 32. For the images per second (imgs/s) metric, higher is better. \label{table_throughput_rebuttal}}
% \vspace{-5mm}
\end{table}

\textbf{Throughput.} In Tab.~\ref{table_throughput}, we compare our models with other low-latency ViTs~\citep{liu2021swin,liu2022convnet,vasu2023fastvit} in terms of throughput to show the efficiency of FFTs in LiSA.
We measure the model throughputs with their respective official source codes by using an RTX6000 with batch size of 32.
With $224\times224$ input, LiSANet-B outperforms both Swin-B~\citep{liu2021swin} and ConvNext models (ConvNext-B, -L)~\citep{liu2022convnet} with higher accuracy.
In addition, HyLiSANet-B outperforms much larger models (Swin-B with $384\times384$ input, ConvNext-L) in all metrics.
FastViT~\citep{vasu2023fastvit} aims to improve the runtime of conventional ViTs by removing skip connections and applying the reparameterization technique.
Regarding FastViT~\citep{vasu2023fastvit}, our models are better in throughput with higher accuracies (e.g.,~6$^{th}$ vs 8$^{th}$ rows), and HyLiSANet-B outperforms FastViT-MA36 in both metrics.
Note that FastViT did not report the results on the $224\times224$ resolution.

\subsection{Downstream tasks on other domains}

% !TEX root = ../main.tex

\begin{table}[t]
  \centering
 \setlength\tabcolsep{5.2pt}
           \scalebox{0.75}{
    \begin{tabular}[t]{clcccc}
    % \begin{tabular}[t]{C{1.0cm}L{3.9cm}C{1.0cm}C{1.2cm}C{0.8cm}} 
    % \begin{tabular}{L{2.2cm}|C{1.1cm}|C{1.1cm}|C{1.1cm}} 
    \toprule
    type & model & pretrain & frame$\times$ & FLOPs  & top-1  \\ 
         &       &          & crops$\times$clips &   & \\
    \midrule    
    \midrule
    % ResNet-18~\citep{he2016deep} & 12    & 1.8    &  69.8 & 89.1  \\
    % % %  RegNetY-1.6GF~\citep{regnet} & 11    & 1.6      & 78.0  & - \\
    % PVT-Ti~\citep{liu2021pay} & 13     & 1.9     & 75.1  & - \\
    % % %  Deit-Ti~\citep{touvron2021training} & --    & ---   &  -- & ---  \\
    % GFNet-H-Ti~\citep{rao2021global} & 15 & 2.1 & 80.1 & 95.1 \\\midrule
    % Ours-H-Ti & 17 & 2.5 & xx.x & xx.x \\\midrule
     \multirow{2}{*}{CNN} & X3D-XL~\citep{feichtenhofer2020x3d}&  - & 16$\times$3$\times$10 & 1452 G & 79.1    \\
     & SlowFast+NL~\citep{feichtenhofer2019slowfast}&  - & 16$\times$3$\times$10 & 7020 G & 79.8    \\ 
     \midrule
     & X-ViT~\citep{bulat2021space}&  IN-21K & 16$\times$3$\times$1 & 850 G & 80.2    \\
     & Mformer-L~\citep{patrick2021keeping}&  IN-21K & 16$\times$3$\times$10 & 35553 G & 80.2    \\
     ViT & ViViT-L~\citep{arnab2021vivit}&  IN-21K & 16$\times$3$\times$4 & 17352 G & 80.6    \\
        & Swin-B~\citep{liu2022video}&  IN-1K & 32$\times$3$\times$4 & 3384 G & 80.6    \\
        & TimeSformer-L~\citep{bertasius2021space}&  IN-21K & 16$\times$3$\times$1 & 7140 G & 80.7    \\
     \midrule
     & MViT-B~\citep{fan2021multiscale}&  - & 16$\times$1$\times$5 & 353 G & 78.4    \\
     \multirow{2}{*}{Hybrid}& UniFormer-S~\citep{li2023uniformer}&  IN-1K & 16$\times$1$\times$4 & 167 G & 80.8    \\
      & MViTv2-S~\citep{li2022mvitv2}&  - & 16$\times$1$\times$5 & 320 G & 81.0    \\
     & UniFormer-B~\citep{li2023uniformer}&  IN-1K & 16$\times$1$\times$4 & 389 G & 82.0    \\
     % \midrule
    %  \bottomrule
 \rowcolor{Gray}
     &HyLiSANet-S &  IN-1K & 16$\times$1$\times$4 & 165 G & 81.1     \\  \rowcolor{Gray}
     &HyLiSANet-B  &  IN-1K & 16$\times$1$\times$4 & 428 G & \textbf{82.5}      \\
     \midrule
    \end{tabular}%
    }
    \caption{\textbf{Comparison to the other models on Kinetics-400.} Pre-trained weights, FLOPs (G), and top-1 accuracy (\%) on Kinetics-400 validation set are shown. All the models use 16 input frames except for Swin~\citep{liu2022video}.} %\vspace{5pt}
  \label{tab:kinetics}
% \vspace{-0.3cm}
%   \vspace{-10pt}
% \afterTable
\end{table}%

\textbf{Video action recognition.}
We conduct experiments on Kinetics-400 by adjusting HyLiSANet-S \& -B for video representation learning. Depthwise convolutions in the models are transformed from 2D to 3D, and LiSA takes ($T\times H\times W$) tokens for spatio-temporal modeling.
We temporally downsample at the first patch embedding layer, and keep the temporal dimension for the rest of the models.
For training, we fine-tune the ImageNet-1K trained weights by inflating 2D convolution kernels and LiSA weights.
We follow the training recipe of Uniformer~\citep{li2023uniformer} and all the experimental details are in Sec.A.2 of the supplementary material.
Table~\ref{tab:kinetics} demonstrates the state-of-the-art results on Kinetics-400 including 3D CNNs~\citep{feichtenhofer2019slowfast,feichtenhofer2020x3d}, video transformers~\citep{arnab2021vivit,patrick2021keeping,liu2022video,bertasius2021space,bulat2021space}, and hybrid models~\citep{li2022mvitv2,fan2021multiscale,li2023uniformer}. For a fair comparison, we compare the models taking the same number of input frames (16 frames).
Our small model, HyLiSANet-S, outperforms CNN-based models~\citep{feichtenhofer2019slowfast,feichtenhofer2020x3d} and ViT models~\citep{arnab2021vivit,patrick2021keeping,liu2022video,bertasius2021space,bulat2021space} in accuracy while consuming much fewer FLOPs. Our base model, HyLiSANet-B, outperforms all the other models including hybrid ViTs~\citep{fan2021multiscale,li2022mvitv2,li2023uniformer} in accuracy with comparable FLOPs.
The results demonstrate the transferability of our models, and further verify that LiSA is highly beneficial for spatio-temporal modeling.

\begin{table}[t]
    \centering
        \captionsetup{width=0.98\columnwidth}
            \centering
            % \begin{tabular}[t]{L{2.2cm}|C{1.1cm}|C{1.1cm}|C{1.1cm}} 
 \setlength\tabcolsep{2.8pt}
            \scalebox{0.75}{
            \begin{tabular}[t]{lcccccccc} 
    % \begin{tabular}[t]{L{2.1cm}C{1.0cm}C{1.2cm}C{0.7cm}C{0.75cm}C{0.75cm}C{0.75cm}C{0.75cm}C{0.75cm}} 
            \toprule
            model &  FLOPs & \#params & \multicolumn{6}{c}{Mask R-CNN 1$\times$ schedule} \\
            % \midrule
             &   &  & AP$^b$ & AP$^b_{50}$ & AP$^b_{75}$ & AP$^m$& AP$^m_{50}$& AP$^m_{75}$ \\
            \midrule
            \midrule
            Res-50~\citep{he2016deep}    & 260 G & 44 M & 38.0 & 58.6 & 41.4 & 34.4 & 55.1 & 36.7 \\
            PVT-S~\citep{wang2021pyramid}  & 245 G & 44 M & 42.9 & 65.8 & 47.1 & 40.0 & 62.7 & 42.9\\
            Twins-S~\citep{chu2021twins}  & 238 G & 44 M & 24.0 & 50.0 & 41.4 & 34.4 & 55.1 & 36.7 \\
            Swin-T~\citep{liu2021swin}  & 264 G & 48 M & 42.2 & 64.6 & 46.2 & 39.1 & 61.6 & 42.0\\
            ViL-S~\citep{zhang2021multi}     & 218 G & 45 M & 44.9 & 67.1 & 49.3 & 41.0 & 64.2 & 44.1\\
            Focal-T~\citep{yang2021focal}   & 291 G & 49 M & 44.8 & 67.7 & 49.2 & 41.0 & 64.7 & 44.2\\
            iFormer-S~\citep{siinception}   & 263 G & 40 M & 46.2 & 68.5 & 50.6 &41.9 & 65.3 & 45.0\\
            CSwin-T~\citep{dong2022cswin}   & 279 G & 42 M & 46.7 & 68.6 & 51.3 &42.2 & 65.6 & 45.4\\
            \midrule
                 \rowcolor{Gray}
            % LiSANet-S & 244 G & 41 M & 46.5 & 67.9 & 50.8 & 41.7 & 65.0 & 44.7 \\  
            LiSANet-S & 258 G & 43 M & 47.2 & 68.3 & 52.0 & 42.2 & 65.7 & 45.2 \\
                 \rowcolor{Gray}
            % HyLiSANet-S & 263 G & 41 M & \textbf{47.2} & \textbf{68.6} & \textbf{51.7} & \textbf{42.4} & \textbf{65.8} & \textbf{45.7} \\
            HyLiSANet-S & 265 G & 42 M & \textbf{47.7} & \textbf{69.0} & \textbf{52.5} & \textbf{42.6} & \textbf{65.9} & \textbf{45.9} \\  
            \midrule
            \end{tabular}
            }
            \caption{\textbf{Comparison with other models on COCO validation set.} FLOPs (G), the number of parameters (M), box mAP (AP$^b$) and mask mAP (AP$^m$) are shown. Note that FLOPs are measured at resolution $800\times 1280$.}
            \label{table_detection}
            % \vspace{-0.3cm}

     % \afterTable
\end{table}

\textbf{Object detection \& instance segmentation.}
To show the generalization ability of LiSA, we conduct object detection experiments on the COCO dataset.
We adopt standard Mask R-CNN~\citep{he2017mask} detection frameworks, which employ ImageNet-1K pre-trained
weights for fine-tuning.
We use a 1$\times$ schedule (12 epochs) and follow the same recipe as in~\citep{liu2021swin}.
In Tab.~\ref{table_detection}, we show the results of object detection and instance segmentation tasks.
Our HyLiSANet-S shows the best performances among CNN~\citep{he2016deep}, ViT~\citep{wang2021pyramid,liu2021swin,zhang2021multi,yang2021focal}, and Hybrid ViT~\citep{siinception} backbones in AP$^b$ and AP$^m$, while maintaining its efficiency.
Since these are high-resolution computer vision tasks (\eg, $800\times1280$), the results demonstrate that LiSA is a proper fit for processing a large number of tokens compared to other attention methods~\citep{wang2021pyramid,liu2021swin,yang2021focal}.

\begin{table}[t]
    \centering
        \captionsetup{width=0.98\columnwidth}
            \centering
            % \begin{tabular}[t]{L{2.2cm}|C{1.1cm}|C{1.1cm}|C{1.1cm}} 
            \scalebox{0.76}{
            % \begin{tabular}[t]{lccc} 
    \begin{tabular}[t]{L{4.2cm}C{1.2cm}C{1.2cm}C{1.5cm}} 
            \toprule
            model &  FLOPs & \#params & mIOU (\%) \\
            \midrule
            \midrule
            ResNet-50~\citep{he2016deep} & 183 G & 29 M & 36.7 \\ 
            PVT-S~\citep{wang2022pvt} & 161 G & 28 M & 39.8 \\
            Twins-S~\citep{chu2021twins} & 144 G & 28 M & 43.2 \\
            Swin-T~\citep{liu2021swin} & 182 G & 32 M & 41.5 \\
            UniFormer-S$_{h32}$~\citep{li2023uniformer} & 199 G & 25 M & 46.2 \\
            UniFormer-S~\citep{li2023uniformer} & 147 G & 25 M  & 46.6 \\
            CSwin-T~\citep{dong2022cswin} & 202 G & 26 M	& 48.2 \\
            \midrule
            \rowcolor{Gray}
            % LiSANet-S (ours)  & 161 G & 25 M & 48.0 \\ 
            LiSANet-S  & 176 G & 27 M & 49.2 \\ 
            \rowcolor{Gray}
            % HyLiSANet-S (ours)  & 182 G & 25 M & \textbf{48.4} \\
            HyLiSANet-S  & 184 G & 26 M & \textbf{49.3} \\
            \midrule
            \end{tabular}
            }
            % \vspace{-0.3cm}
            \caption{\textbf{Comparison with other models on ADE-20K}. mIOU (\%), FLOPs (G) and the number of parameters (M) are shown. Note that FLOPs are measured at resolution 512$\times$2048.}
            \label{table_semantic_segmentation}
     % \afterTable

    % \vspace{-0.4cm}
\end{table}

\textbf{Semantic segmentation.} We also evaluate our model on ADE-20K dataset~\citep{zhou2017ade20k}. We adopt the popular Semantic FPN~\citep{kirillov2019panoptic} as a basic framework and the model is trained for 80k iterations. The stochastic depth rate is set as 0.15, and we follow the same setting of PVT~\citep{wang2022pvt} for a fair comparison. Table~\ref{table_semantic_segmentation} summarizes the results on ADE-20K. 
% LiSANet-S is quite competitive to the other models with lower computation and 
HyLiSANet-S achieves the best mIOU among different models while requiring fewer FLOPs and the number of parameters, indicating that LiSA is beneficial for processing high-resolutions.

\begin{figure}[t]
    \includegraphics[width=\columnwidth]{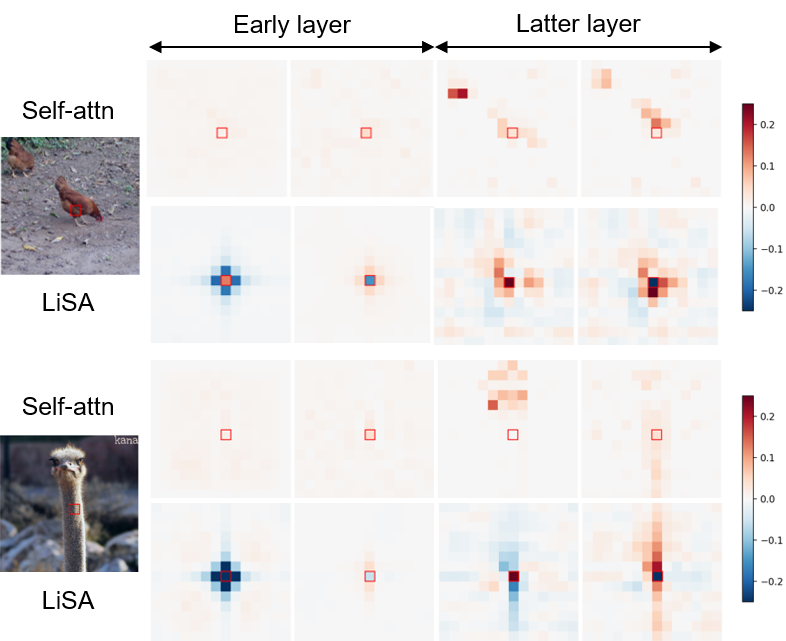}
    \caption{
    \textbf{Attention kernels of self-attention \& LiSA.} Attention kernels from different layers and heads are visualized. For each sample, the top row is self-attention, and the bottom is LiSA. Note that the red box in the center of each subfigure is the query pixel.
    }
    \label{fig:viz_attn_map}
    % \afterCaption
    % \vspace{-0.5cm}
\end{figure}
% hen cls_num=8
% ostrich cls_num=9
% robin cls_num = 15

\subsection{Visualization}
In Fig.~\ref{fig:viz_attn_map}, we visualize both self-attention and LiSA kernels of different layers and heads from isotropic models.
As expected, LiSA kernels contain much more diverse patterns compared to self-attention kernels.
Self-attention kernels in the early layers often fail to capture relevant context, and those in the latter layers are effective, but they usually capture redundant information.
Unlike self-attention, LiSA kernels in the early layers focus on encoding local features.
Some of these look similar to Sobel or Laplacian filters, which are beneficial for learning local structural information.
Considering that modern hybrid models~\citep{xiao2021early,dai2021coatnet,li2022uniformer,siinception}, which replace self-attention with convolution in early layers obtain an extra accuracy gain, the behavior of LiSA kernels in early layers seems reasonable.
Meanwhile, LiSA kernels in the latter layers concentrate on the context relevant to the target object like self-attention does.
LiSA, however, generates more diverse shapes of kernels than self-attention, which implies that they aggregate the relevant context and further consider structural patterns inside the context at the aggregation.
Therefore, this visualization demonstrates that our structure-aware attention kernel can be more expressive and flexible compared to the self-attention kernel.

\section{Conclusion}

In this paper, we have presented LiSA, a novel expressive, yet efficient, attention operator that learns rich structural patterns with log-linear complexity. Our comprehensive analyses have shown that the ViTs based on LiSA, LiSANets, outperform their counterparts in accuracy and computational complexity. LiSANet \& HyLiSANet have achieved competitive performance on various kinds of visual understanding tasks such as image classification, video classification, object detection, and semantic segmentation. While LiSA is effective yet efficient for understanding visual concepts, it still leaves room for improvement in several aspects. First, LiSA encodes structural patterns as a fixed size of the vector, which is set as a parameter $D$, but we can further investigate a method to dynamically change the size of structural patterns depending on the visual context. Second, although LiSA is much more efficient than the other attention methods due to the log-linear complexity, it is still heavy to construct multiple layers for processing high resolutions. It would be interesting to examine the effect of LiSA with the local or sparse attention techniques.
We believe that LiSA can be a guideline for designing a better basic brick for visual understanding, and we further expect our structure-aware attention could be applied for cross-attention mechanisms, which are widely used for handling multiple modalities such as image-text or video-text.

\section*{Acknowledgments}
This work has been supported in part by the ANR grant AVENUE
(ANR-18-CE23-0011), the Junta de Andaluc\'ia of Spain (P18-FR-3130 and P20\_00430, including European Union funds) and the Ministry of Education of Spain (PID2019-105396RB-I00). We also thank the EuroHPC JU for the GPU computing hours.

% \bibliography{sn-bibliography}% common bib file
\bibliography{HKwon2,local}% common bib file


\begin{thebibliography}{85}
\ifx \bisbn   \undefined \def \bisbn  #1{ISBN #1}\fi
\ifx \binits  \undefined \def \binits#1{#1}\fi
\ifx \bauthor  \undefined \def \bauthor#1{#1}\fi
\ifx \batitle  \undefined \def \batitle#1{#1}\fi
\ifx \bjtitle  \undefined \def \bjtitle#1{#1}\fi
\ifx \bvolume  \undefined \def \bvolume#1{\textbf{#1}}\fi
\ifx \byear  \undefined \def \byear#1{#1}\fi
\ifx \bissue  \undefined \def \bissue#1{#1}\fi
\ifx \bfpage  \undefined \def \bfpage#1{#1}\fi
\ifx \blpage  \undefined \def \blpage #1{#1}\fi
\ifx \burl  \undefined \def \burl#1{\textsf{#1}}\fi
\ifx \doiurl  \undefined \def \doiurl#1{\url{https://doi.org/#1}}\fi
\ifx \betal  \undefined \def \betal{\textit{et al.}}\fi
\ifx \binstitute  \undefined \def \binstitute#1{#1}\fi
\ifx \binstitutionaled  \undefined \def \binstitutionaled#1{#1}\fi
\ifx \bctitle  \undefined \def \bctitle#1{#1}\fi
\ifx \beditor  \undefined \def \beditor#1{#1}\fi
\ifx \bpublisher  \undefined \def \bpublisher#1{#1}\fi
\ifx \bbtitle  \undefined \def \bbtitle#1{#1}\fi
\ifx \bedition  \undefined \def \bedition#1{#1}\fi
\ifx \bseriesno  \undefined \def \bseriesno#1{#1}\fi
\ifx \blocation  \undefined \def \blocation#1{#1}\fi
\ifx \bsertitle  \undefined \def \bsertitle#1{#1}\fi
\ifx \bsnm \undefined \def \bsnm#1{#1}\fi
\ifx \bsuffix \undefined \def \bsuffix#1{#1}\fi
\ifx \bparticle \undefined \def \bparticle#1{#1}\fi
\ifx \barticle \undefined \def \barticle#1{#1}\fi
\bibcommenthead
\ifx \bconfdate \undefined \def \bconfdate #1{#1}\fi
\ifx \botherref \undefined \def \botherref #1{#1}\fi
\ifx \url \undefined \def \url#1{\textsf{#1}}\fi
\ifx \bchapter \undefined \def \bchapter#1{#1}\fi
\ifx \bbook \undefined \def \bbook#1{#1}\fi
\ifx \bcomment \undefined \def \bcomment#1{#1}\fi
\ifx \oauthor \undefined \def \oauthor#1{#1}\fi
\ifx \citeauthoryear \undefined \def \citeauthoryear#1{#1}\fi
\ifx \endbibitem  \undefined \def \endbibitem {}\fi
\ifx \bconflocation  \undefined \def \bconflocation#1{#1}\fi
\ifx \arxivurl  \undefined \def \arxivurl#1{\textsf{#1}}\fi
\csname PreBibitemsHook\endcsname

%%% 1
\bibitem[\protect\citeauthoryear{Dosovitskiy
  et~al.}{2020}]{dosovitskiy2020image}
\begin{botherref}
\oauthor{\bsnm{Dosovitskiy}, \binits{A.}},
\oauthor{\bsnm{Beyer}, \binits{L.}},
\oauthor{\bsnm{Kolesnikov}, \binits{A.}},
\oauthor{\bsnm{Weissenborn}, \binits{D.}},
\oauthor{\bsnm{Zhai}, \binits{X.}},
\oauthor{\bsnm{Unterthiner}, \binits{T.}},
\oauthor{\bsnm{Dehghani}, \binits{M.}},
\oauthor{\bsnm{Minderer}, \binits{M.}},
\oauthor{\bsnm{Heigold}, \binits{G.}},
\oauthor{\bsnm{Gelly}, \binits{S.}}, et al.:
An image is worth 16x16 words: Transformers for image recognition at scale.
\textit{Proc. International Conference on Learning Representations (ICLR)}
(2020)
\end{botherref}
\endbibitem

%%% 2
\bibitem[\protect\citeauthoryear{Li et~al.}{2022}]{li2022uniformer}
\begin{botherref}
\oauthor{\bsnm{Li}, \binits{K.}},
\oauthor{\bsnm{Wang}, \binits{Y.}},
\oauthor{\bsnm{Gao}, \binits{P.}},
\oauthor{\bsnm{Song}, \binits{G.}},
\oauthor{\bsnm{Liu}, \binits{Y.}},
\oauthor{\bsnm{Li}, \binits{H.}},
\oauthor{\bsnm{Qiao}, \binits{Y.}}:
Uniformer: Unified transformer for efficient spatiotemporal representation
  learning.
\textit{Proc. International Conference on Learning Representations (ICLR)}
(2022)
\end{botherref}
\endbibitem

%%% 3
\bibitem[\protect\citeauthoryear{Yuan et~al.}{2021}]{yuan2021tokens}
\begin{bchapter}
\bauthor{\bsnm{Yuan}, \binits{L.}},
\bauthor{\bsnm{Chen}, \binits{Y.}},
\bauthor{\bsnm{Wang}, \binits{T.}},
\bauthor{\bsnm{Yu}, \binits{W.}},
\bauthor{\bsnm{Shi}, \binits{Y.}},
\bauthor{\bsnm{Jiang}, \binits{Z.-H.}},
\bauthor{\bsnm{Tay}, \binits{F.E.}},
\bauthor{\bsnm{Feng}, \binits{J.}},
\bauthor{\bsnm{Yan}, \binits{S.}}:
\bctitle{Tokens-to-token vit: Training vision transformers from scratch on
  imagenet}.
In: \bbtitle{\textit{Proc. IEEE International Conference on Computer Vision
  (ICCV)}},
pp. \bfpage{558}--\blpage{567}
(\byear{2021})
\end{bchapter}
\endbibitem

%%% 4
\bibitem[\protect\citeauthoryear{Touvron et~al.}{2021}]{touvron2021training}
\begin{bchapter}
\bauthor{\bsnm{Touvron}, \binits{H.}},
\bauthor{\bsnm{Cord}, \binits{M.}},
\bauthor{\bsnm{Douze}, \binits{M.}},
\bauthor{\bsnm{Massa}, \binits{F.}},
\bauthor{\bsnm{Sablayrolles}, \binits{A.}},
\bauthor{\bsnm{J{\'e}gou}, \binits{H.}}:
\bctitle{Training data-efficient image transformers \& distillation through
  attention}.
In: \bbtitle{\textit{Proc. International Conference on Machine Learning
  (ICML)}},
pp. \bfpage{10347}--\blpage{10357}
(\byear{2021}).
\bcomment{PMLR}
\end{bchapter}
\endbibitem

%%% 5
\bibitem[\protect\citeauthoryear{Raffel et~al.}{2019}]{raffel2019exploring}
\begin{botherref}
\oauthor{\bsnm{Raffel}, \binits{C.}},
\oauthor{\bsnm{Shazeer}, \binits{N.}},
\oauthor{\bsnm{Roberts}, \binits{A.}},
\oauthor{\bsnm{Lee}, \binits{K.}},
\oauthor{\bsnm{Narang}, \binits{S.}},
\oauthor{\bsnm{Matena}, \binits{M.}},
\oauthor{\bsnm{Zhou}, \binits{Y.}},
\oauthor{\bsnm{Li}, \binits{W.}},
\oauthor{\bsnm{Liu}, \binits{P.J.}}:
Exploring the limits of transfer learning with a unified text-to-text
  transformer.
\emph{arXiv preprint arXiv:1910.10683}
(2019)
\end{botherref}
\endbibitem

%%% 6
\bibitem[\protect\citeauthoryear{Liu et~al.}{2021}]{liu2021swin}
\begin{bchapter}
\bauthor{\bsnm{Liu}, \binits{Z.}},
\bauthor{\bsnm{Lin}, \binits{Y.}},
\bauthor{\bsnm{Cao}, \binits{Y.}},
\bauthor{\bsnm{Hu}, \binits{H.}},
\bauthor{\bsnm{Wei}, \binits{Y.}},
\bauthor{\bsnm{Zhang}, \binits{Z.}},
\bauthor{\bsnm{Lin}, \binits{S.}},
\bauthor{\bsnm{Guo}, \binits{B.}}:
\bctitle{Swin transformer: Hierarchical vision transformer using shifted
  windows}.
In: \bbtitle{\textit{Proc. IEEE International Conference on Computer Vision
  (ICCV)}},
pp. \bfpage{10012}--\blpage{10022}
(\byear{2021})
\end{bchapter}
\endbibitem

%%% 7
\bibitem[\protect\citeauthoryear{Bello}{2020}]{bello2021lambdanetworks}
\begin{bchapter}
\bauthor{\bsnm{Bello}, \binits{I.}}:
\bctitle{Lambdanetworks: Modeling long-range interactions without attention}.
In: \bbtitle{\textit{Proc. International Conference on Learning Representations
  (ICLR)}}
(\byear{2020})
\end{bchapter}
\endbibitem

%%% 8
\bibitem[\protect\citeauthoryear{Zhao et~al.}{2020}]{zhao2020exploring}
\begin{bchapter}
\bauthor{\bsnm{Zhao}, \binits{H.}},
\bauthor{\bsnm{Jia}, \binits{J.}},
\bauthor{\bsnm{Koltun}, \binits{V.}}:
\bctitle{Exploring self-attention for image recognition}.
In: \bbtitle{\textit{Proc. IEEE Conference on Computer Vision and Pattern
  Recognition (CVPR)}},
pp. \bfpage{10076}--\blpage{10085}
(\byear{2020})
\end{bchapter}
\endbibitem

%%% 9
\bibitem[\protect\citeauthoryear{Kim et~al.}{2021}]{kim2021relational}
\begin{barticle}
\bauthor{\bsnm{Kim}, \binits{M.}},
\bauthor{\bsnm{Kwon}, \binits{H.}},
\bauthor{\bsnm{Wang}, \binits{C.}},
\bauthor{\bsnm{Kwak}, \binits{S.}},
\bauthor{\bsnm{Cho}, \binits{M.}}:
\batitle{Relational self-attention: What's missing in attention for video
  understanding}.
\bjtitle{\textit{Proc. Neural Information Processing Systems (NeurIPS)}}
\bvolume{34},
\bfpage{8046}--\blpage{8059}
(\byear{2021})
\end{barticle}
\endbibitem

%%% 10
\bibitem[\protect\citeauthoryear{Wang et~al.}{2020}]{wang2020linformer}
\begin{botherref}
\oauthor{\bsnm{Wang}, \binits{S.}},
\oauthor{\bsnm{Li}, \binits{B.}},
\oauthor{\bsnm{Khabsa}, \binits{M.}},
\oauthor{\bsnm{Fang}, \binits{H.}},
\oauthor{\bsnm{Ma}, \binits{H.}}:
Linformer: Self-attention with linear complexity.
\emph{arXiv preprint arXiv:2006.04768}
(2020)
\end{botherref}
\endbibitem

%%% 11
\bibitem[\protect\citeauthoryear{Choromanski
  et~al.}{2021}]{choromanski2020rethinking}
\begin{bchapter}
\bauthor{\bsnm{Choromanski}, \binits{K.}},
\bauthor{\bsnm{Likhosherstov}, \binits{V.}},
\bauthor{\bsnm{Dohan}, \binits{D.}},
\bauthor{\bsnm{Song}, \binits{X.}},
\bauthor{\bsnm{Gane}, \binits{A.}},
\bauthor{\bsnm{Sarlos}, \binits{T.}},
\bauthor{\bsnm{Hawkins}, \binits{P.}},
\bauthor{\bsnm{Davis}, \binits{J.}},
\bauthor{\bsnm{Mohiuddin}, \binits{A.}},
\bauthor{\bsnm{Kaiser}, \binits{L.}}, \betal:
\bctitle{Rethinking attention with performers}.
In: \bbtitle{\textit{Proc. International Conference on Learning Representations
  (ICLR)}}
(\byear{2021})
\end{bchapter}
\endbibitem

%%% 12
\bibitem[\protect\citeauthoryear{Qin et~al.}{2022}]{qin2022cosformer}
\begin{botherref}
\oauthor{\bsnm{Qin}, \binits{Z.}},
\oauthor{\bsnm{Sun}, \binits{W.}},
\oauthor{\bsnm{Deng}, \binits{H.}},
\oauthor{\bsnm{Li}, \binits{D.}},
\oauthor{\bsnm{Wei}, \binits{Y.}},
\oauthor{\bsnm{Lv}, \binits{B.}},
\oauthor{\bsnm{Yan}, \binits{J.}},
\oauthor{\bsnm{Kong}, \binits{L.}},
\oauthor{\bsnm{Zhong}, \binits{Y.}}:
cosformer: Rethinking softmax in attention.
\emph{arXiv preprint arXiv:2202.08791}
(2022)
\end{botherref}
\endbibitem

%%% 13
\bibitem[\protect\citeauthoryear{Liutkus et~al.}{2021}]{liutkus2021relative}
\begin{bchapter}
\bauthor{\bsnm{Liutkus}, \binits{A.}},
\bauthor{\bsnm{C{\i}fka}, \binits{O.}},
\bauthor{\bsnm{Wu}, \binits{S.-L.}},
\bauthor{\bsnm{Simsekli}, \binits{U.}},
\bauthor{\bsnm{Yang}, \binits{Y.-H.}},
\bauthor{\bsnm{Richard}, \binits{G.}}:
\bctitle{Relative positional encoding for transformers with linear complexity}.
In: \bbtitle{\textit{Proc. International Conference on Machine Learning
  (ICML)}},
pp. \bfpage{7067}--\blpage{7079}
(\byear{2021}).
\bcomment{PMLR}
\end{bchapter}
\endbibitem

%%% 14
\bibitem[\protect\citeauthoryear{Chen}{2021}]{chen2021permuteformer}
\begin{bchapter}
\bauthor{\bsnm{Chen}, \binits{P.}}:
\bctitle{Permuteformer: Efficient relative position encoding for long
  sequences}.
In: \bbtitle{\emph{Proceedings of the 2021 Conference on Empirical Methods in
  Natural Language Processing}},
pp. \bfpage{10606}--\blpage{10618}
(\byear{2021})
\end{bchapter}
\endbibitem

%%% 15
\bibitem[\protect\citeauthoryear{Luo et~al.}{2021}]{luo2021stable}
\begin{barticle}
\bauthor{\bsnm{Luo}, \binits{S.}},
\bauthor{\bsnm{Li}, \binits{S.}},
\bauthor{\bsnm{Cai}, \binits{T.}},
\bauthor{\bsnm{He}, \binits{D.}},
\bauthor{\bsnm{Peng}, \binits{D.}},
\bauthor{\bsnm{Zheng}, \binits{S.}},
\bauthor{\bsnm{Ke}, \binits{G.}},
\bauthor{\bsnm{Wang}, \binits{L.}},
\bauthor{\bsnm{Liu}, \binits{T.-Y.}}:
\batitle{Stable, fast and accurate: Kernelized attention with relative
  positional encoding}.
\bjtitle{\textit{Proc. Neural Information Processing Systems (NeurIPS)}}
\bvolume{34},
\bfpage{22795}--\blpage{22807}
(\byear{2021})
\end{barticle}
\endbibitem

%%% 16
\bibitem[\protect\citeauthoryear{Li et~al.}{2022}]{li2022mvitv2}
\begin{bchapter}
\bauthor{\bsnm{Li}, \binits{Y.}},
\bauthor{\bsnm{Wu}, \binits{C.-Y.}},
\bauthor{\bsnm{Fan}, \binits{H.}},
\bauthor{\bsnm{Mangalam}, \binits{K.}},
\bauthor{\bsnm{Xiong}, \binits{B.}},
\bauthor{\bsnm{Malik}, \binits{J.}},
\bauthor{\bsnm{Feichtenhofer}, \binits{C.}}:
\bctitle{Mvitv2: Improved multiscale vision transformers for classification and
  detection}.
In: \bbtitle{\textit{Proc. IEEE Conference on Computer Vision and Pattern
  Recognition (CVPR)}},
pp. \bfpage{4804}--\blpage{4814}
(\byear{2022})
\end{bchapter}
\endbibitem

%%% 17
\bibitem[\protect\citeauthoryear{Wu et~al.}{2021}]{wu2021cvt}
\begin{bchapter}
\bauthor{\bsnm{Wu}, \binits{H.}},
\bauthor{\bsnm{Xiao}, \binits{B.}},
\bauthor{\bsnm{Codella}, \binits{N.}},
\bauthor{\bsnm{Liu}, \binits{M.}},
\bauthor{\bsnm{Dai}, \binits{X.}},
\bauthor{\bsnm{Yuan}, \binits{L.}},
\bauthor{\bsnm{Zhang}, \binits{L.}}:
\bctitle{Cvt: Introducing convolutions to vision transformers}.
In: \bbtitle{\textit{Proc. IEEE International Conference on Computer Vision
  (ICCV)}},
pp. \bfpage{22}--\blpage{31}
(\byear{2021})
\end{bchapter}
\endbibitem

%%% 18
\bibitem[\protect\citeauthoryear{Wang et~al.}{2022}]{wang2022pvt}
\begin{barticle}
\bauthor{\bsnm{Wang}, \binits{W.}},
\bauthor{\bsnm{Xie}, \binits{E.}},
\bauthor{\bsnm{Li}, \binits{X.}},
\bauthor{\bsnm{Fan}, \binits{D.-P.}},
\bauthor{\bsnm{Song}, \binits{K.}},
\bauthor{\bsnm{Liang}, \binits{D.}},
\bauthor{\bsnm{Lu}, \binits{T.}},
\bauthor{\bsnm{Luo}, \binits{P.}},
\bauthor{\bsnm{Shao}, \binits{L.}}:
\batitle{Pvt v2: Improved baselines with pyramid vision transformer}.
\bjtitle{\emph{Computational Visual Media}}
\bvolume{8}(\bissue{3}),
\bfpage{415}--\blpage{424}
(\byear{2022})
\end{barticle}
\endbibitem

%%% 19
\bibitem[\protect\citeauthoryear{Si et~al.}{2022}]{siinception}
\begin{barticle}
\bauthor{\bsnm{Si}, \binits{C.}},
\bauthor{\bsnm{Yu}, \binits{W.}},
\bauthor{\bsnm{Zhou}, \binits{P.}},
\bauthor{\bsnm{Zhou}, \binits{Y.}},
\bauthor{\bsnm{Wang}, \binits{X.}},
\bauthor{\bsnm{Yan}, \binits{S.}}:
\batitle{Inception transformer}.
\bjtitle{\textit{Proc. Neural Information Processing Systems (NeurIPS)}}
\bvolume{35},
\bfpage{23495}--\blpage{23509}
(\byear{2022})
\end{barticle}
\endbibitem

%%% 20
\bibitem[\protect\citeauthoryear{Tu et~al.}{2022}]{tu2022maxvit}
\begin{bchapter}
\bauthor{\bsnm{Tu}, \binits{Z.}},
\bauthor{\bsnm{Talebi}, \binits{H.}},
\bauthor{\bsnm{Zhang}, \binits{H.}},
\bauthor{\bsnm{Yang}, \binits{F.}},
\bauthor{\bsnm{Milanfar}, \binits{P.}},
\bauthor{\bsnm{Bovik}, \binits{A.}},
\bauthor{\bsnm{Li}, \binits{Y.}}:
\bctitle{Maxvit: Multi-axis vision transformer}.
In: \bbtitle{\textit{Proc. European Conference on Computer Vision (ECCV)}},
pp. \bfpage{459}--\blpage{479}
(\byear{2022}).
\bcomment{Springer}
\end{bchapter}
\endbibitem

%%% 21
\bibitem[\protect\citeauthoryear{Shechtman and
  Irani}{2007}]{shechtman2007matching}
\begin{bchapter}
\bauthor{\bsnm{Shechtman}, \binits{E.}},
\bauthor{\bsnm{Irani}, \binits{M.}}:
\bctitle{Matching local self-similarities across images and videos}.
In: \bbtitle{\textit{Proc. IEEE Conference on Computer Vision and Pattern
  Recognition (CVPR)}},
pp. \bfpage{1}--\blpage{8}
(\byear{2007}).
\bcomment{IEEE}
\end{bchapter}
\endbibitem

%%% 22
\bibitem[\protect\citeauthoryear{Wang et~al.}{2020}]{wang2020video}
\begin{bchapter}
\bauthor{\bsnm{Wang}, \binits{H.}},
\bauthor{\bsnm{Tran}, \binits{D.}},
\bauthor{\bsnm{Torresani}, \binits{L.}},
\bauthor{\bsnm{Feiszli}, \binits{M.}}:
\bctitle{Video modeling with correlation networks}.
In: \bbtitle{\textit{Proc. IEEE Conference on Computer Vision and Pattern
  Recognition (CVPR)}},
pp. \bfpage{352}--\blpage{361}
(\byear{2020})
\end{bchapter}
\endbibitem

%%% 23
\bibitem[\protect\citeauthoryear{Kwon et~al.}{2021}]{kwon2021learning}
\begin{bchapter}
\bauthor{\bsnm{Kwon}, \binits{H.}},
\bauthor{\bsnm{Kim}, \binits{M.}},
\bauthor{\bsnm{Kwak}, \binits{S.}},
\bauthor{\bsnm{Cho}, \binits{M.}}:
\bctitle{Learning self-similarity in space and time as generalized motion for
  video action recognition}.
In: \bbtitle{\textit{Proc. IEEE International Conference on Computer Vision
  (ICCV)}},
pp. \bfpage{13065}--\blpage{13075}
(\byear{2021})
\end{bchapter}
\endbibitem

%%% 24
\bibitem[\protect\citeauthoryear{Sun et~al.}{2018}]{sun2018pwc}
\begin{bchapter}
\bauthor{\bsnm{Sun}, \binits{D.}},
\bauthor{\bsnm{Yang}, \binits{X.}},
\bauthor{\bsnm{Liu}, \binits{M.-Y.}},
\bauthor{\bsnm{Kautz}, \binits{J.}}:
\bctitle{Pwc-net: Cnns for optical flow using pyramid, warping, and cost
  volume}.
In: \bbtitle{\textit{Proc. IEEE Conference on Computer Vision and Pattern
  Recognition (CVPR)}},
pp. \bfpage{8934}--\blpage{8943}
(\byear{2018})
\end{bchapter}
\endbibitem

%%% 25
\bibitem[\protect\citeauthoryear{Deng et~al.}{2009}]{deng2009imagenet}
\begin{bchapter}
\bauthor{\bsnm{Deng}, \binits{J.}},
\bauthor{\bsnm{Dong}, \binits{W.}},
\bauthor{\bsnm{Socher}, \binits{R.}},
\bauthor{\bsnm{Li}, \binits{L.-J.}},
\bauthor{\bsnm{Li}, \binits{K.}},
\bauthor{\bsnm{Fei-Fei}, \binits{L.}}:
\bctitle{Imagenet: A large-scale hierarchical image database}.
In: \bbtitle{\textit{Proc. IEEE Conference on Computer Vision and Pattern
  Recognition (CVPR)}},
pp. \bfpage{248}--\blpage{255}
(\byear{2009}).
\bcomment{IEEE}
\end{bchapter}
\endbibitem

%%% 26
\bibitem[\protect\citeauthoryear{Kay et~al.}{2017}]{kay2017kinetics}
\begin{botherref}
\oauthor{\bsnm{Kay}, \binits{W.}},
\oauthor{\bsnm{Carreira}, \binits{J.}},
\oauthor{\bsnm{Simonyan}, \binits{K.}},
\oauthor{\bsnm{Zhang}, \binits{B.}},
\oauthor{\bsnm{Hillier}, \binits{C.}},
\oauthor{\bsnm{Vijayanarasimhan}, \binits{S.}},
\oauthor{\bsnm{Viola}, \binits{F.}},
\oauthor{\bsnm{Green}, \binits{T.}},
\oauthor{\bsnm{Back}, \binits{T.}},
\oauthor{\bsnm{Natsev}, \binits{P.}}, et al.:
The kinetics human action video dataset.
\emph{arXiv preprint arXiv:1705.06950}
(2017)
\end{botherref}
\endbibitem

%%% 27
\bibitem[\protect\citeauthoryear{Lin et~al.}{2014}]{lin2014microsoft}
\begin{bchapter}
\bauthor{\bsnm{Lin}, \binits{T.-Y.}},
\bauthor{\bsnm{Maire}, \binits{M.}},
\bauthor{\bsnm{Belongie}, \binits{S.}},
\bauthor{\bsnm{Hays}, \binits{J.}},
\bauthor{\bsnm{Perona}, \binits{P.}},
\bauthor{\bsnm{Ramanan}, \binits{D.}},
\bauthor{\bsnm{Doll{\'a}r}, \binits{P.}},
\bauthor{\bsnm{Zitnick}, \binits{C.L.}}:
\bctitle{Microsoft coco: Common objects in context}.
In: \bbtitle{\textit{Proc. European Conference on Computer Vision (ECCV)}},
pp. \bfpage{740}--\blpage{755}
(\byear{2014}).
\bcomment{Springer}
\end{bchapter}
\endbibitem

%%% 28
\bibitem[\protect\citeauthoryear{Zhou et~al.}{2017}]{zhou2017ade20k}
\begin{bchapter}
\bauthor{\bsnm{Zhou}, \binits{B.}},
\bauthor{\bsnm{Zhao}, \binits{H.}},
\bauthor{\bsnm{Puig}, \binits{X.}},
\bauthor{\bsnm{Fidler}, \binits{S.}},
\bauthor{\bsnm{Barriuso}, \binits{A.}},
\bauthor{\bsnm{Torralba}, \binits{A.}}:
\bctitle{Scene parsing through ade20k dataset}.
In: \bbtitle{\textit{Proc. IEEE Conference on Computer Vision and Pattern
  Recognition (CVPR)}},
pp. \bfpage{633}--\blpage{641}
(\byear{2017})
\end{bchapter}
\endbibitem

%%% 29
\bibitem[\protect\citeauthoryear{Carion et~al.}{2020}]{carion2020end}
\begin{bchapter}
\bauthor{\bsnm{Carion}, \binits{N.}},
\bauthor{\bsnm{Massa}, \binits{F.}},
\bauthor{\bsnm{Synnaeve}, \binits{G.}},
\bauthor{\bsnm{Usunier}, \binits{N.}},
\bauthor{\bsnm{Kirillov}, \binits{A.}},
\bauthor{\bsnm{Zagoruyko}, \binits{S.}}:
\bctitle{End-to-end object detection with transformers}.
In: \bbtitle{\textit{Proc. European Conference on Computer Vision (ECCV)}},
pp. \bfpage{213}--\blpage{229}
(\byear{2020}).
\bcomment{Springer}
\end{bchapter}
\endbibitem

%%% 30
\bibitem[\protect\citeauthoryear{Strudel et~al.}{2021}]{strudel2021segmenter}
\begin{bchapter}
\bauthor{\bsnm{Strudel}, \binits{R.}},
\bauthor{\bsnm{Garcia}, \binits{R.}},
\bauthor{\bsnm{Laptev}, \binits{I.}},
\bauthor{\bsnm{Schmid}, \binits{C.}}:
\bctitle{Segmenter: Transformer for semantic segmentation}.
In: \bbtitle{\textit{Proc. IEEE International Conference on Computer Vision
  (ICCV)}},
pp. \bfpage{7262}--\blpage{7272}
(\byear{2021})
\end{bchapter}
\endbibitem

%%% 31
\bibitem[\protect\citeauthoryear{Arnab et~al.}{2021}]{arnab2021vivit}
\begin{bchapter}
\bauthor{\bsnm{Arnab}, \binits{A.}},
\bauthor{\bsnm{Dehghani}, \binits{M.}},
\bauthor{\bsnm{Heigold}, \binits{G.}},
\bauthor{\bsnm{Sun}, \binits{C.}},
\bauthor{\bsnm{Lu{\v{c}}i{\'c}}, \binits{M.}},
\bauthor{\bsnm{Schmid}, \binits{C.}}:
\bctitle{Vivit: A video vision transformer}.
In: \bbtitle{\textit{Proc. IEEE International Conference on Computer Vision
  (ICCV)}},
pp. \bfpage{6836}--\blpage{6846}
(\byear{2021})
\end{bchapter}
\endbibitem

%%% 32
\bibitem[\protect\citeauthoryear{Vaswani et~al.}{2021}]{vaswani2021scaling}
\begin{bchapter}
\bauthor{\bsnm{Vaswani}, \binits{A.}},
\bauthor{\bsnm{Ramachandran}, \binits{P.}},
\bauthor{\bsnm{Srinivas}, \binits{A.}},
\bauthor{\bsnm{Parmar}, \binits{N.}},
\bauthor{\bsnm{Hechtman}, \binits{B.}},
\bauthor{\bsnm{Shlens}, \binits{J.}}:
\bctitle{Scaling local self-attention for parameter efficient visual
  backbones}.
In: \bbtitle{\textit{Proc. IEEE Conference on Computer Vision and Pattern
  Recognition (CVPR)}},
pp. \bfpage{12894}--\blpage{12904}
(\byear{2021})
\end{bchapter}
\endbibitem

%%% 33
\bibitem[\protect\citeauthoryear{Xiao et~al.}{2021}]{xiao2021early}
\begin{barticle}
\bauthor{\bsnm{Xiao}, \binits{T.}},
\bauthor{\bsnm{Singh}, \binits{M.}},
\bauthor{\bsnm{Mintun}, \binits{E.}},
\bauthor{\bsnm{Darrell}, \binits{T.}},
\bauthor{\bsnm{Doll{\'a}r}, \binits{P.}},
\bauthor{\bsnm{Girshick}, \binits{R.}}:
\batitle{Early convolutions help transformers see better}.
\bjtitle{\textit{Proc. Neural Information Processing Systems (NeurIPS)}}
\bvolume{34},
\bfpage{30392}--\blpage{30400}
(\byear{2021})
\end{barticle}
\endbibitem

%%% 34
\bibitem[\protect\citeauthoryear{Han et~al.}{2021}]{han2021transformer}
\begin{barticle}
\bauthor{\bsnm{Han}, \binits{K.}},
\bauthor{\bsnm{Xiao}, \binits{A.}},
\bauthor{\bsnm{Wu}, \binits{E.}},
\bauthor{\bsnm{Guo}, \binits{J.}},
\bauthor{\bsnm{Xu}, \binits{C.}},
\bauthor{\bsnm{Wang}, \binits{Y.}}:
\batitle{Transformer in transformer}.
\bjtitle{\textit{Proc. Neural Information Processing Systems (NeurIPS)}}
\bvolume{34},
\bfpage{15908}--\blpage{15919}
(\byear{2021})
\end{barticle}
\endbibitem

%%% 35
\bibitem[\protect\citeauthoryear{Zhang et~al.}{2022}]{zhang2021aggregating}
\begin{bchapter}
\bauthor{\bsnm{Zhang}, \binits{Z.}},
\bauthor{\bsnm{Zhang}, \binits{H.}},
\bauthor{\bsnm{Zhao}, \binits{L.}},
\bauthor{\bsnm{Chen}, \binits{T.}},
\bauthor{\bsnm{Arik}, \binits{S.{\"O}.}},
\bauthor{\bsnm{Pfister}, \binits{T.}}:
\bctitle{Nested hierarchical transformer: Towards accurate, data-efficient and
  interpretable visual understanding}.
In: \bbtitle{\textit{Proc. AAAI Conference on Artificial Intelligence (AAAI)}},
vol. \bseriesno{36},
pp. \bfpage{3417}--\blpage{3425}
(\byear{2022})
\end{bchapter}
\endbibitem

%%% 36
\bibitem[\protect\citeauthoryear{Dong et~al.}{2022}]{dong2022cswin}
\begin{bchapter}
\bauthor{\bsnm{Dong}, \binits{X.}},
\bauthor{\bsnm{Bao}, \binits{J.}},
\bauthor{\bsnm{Chen}, \binits{D.}},
\bauthor{\bsnm{Zhang}, \binits{W.}},
\bauthor{\bsnm{Yu}, \binits{N.}},
\bauthor{\bsnm{Yuan}, \binits{L.}},
\bauthor{\bsnm{Chen}, \binits{D.}},
\bauthor{\bsnm{Guo}, \binits{B.}}:
\bctitle{Cswin transformer: A general vision transformer backbone with
  cross-shaped windows}.
In: \bbtitle{\textit{Proc. IEEE Conference on Computer Vision and Pattern
  Recognition (CVPR)}},
pp. \bfpage{12124}--\blpage{12134}
(\byear{2022})
\end{bchapter}
\endbibitem

%%% 37
\bibitem[\protect\citeauthoryear{Dai et~al.}{2021}]{dai2021coatnet}
\begin{barticle}
\bauthor{\bsnm{Dai}, \binits{Z.}},
\bauthor{\bsnm{Liu}, \binits{H.}},
\bauthor{\bsnm{Le}, \binits{Q.V.}},
\bauthor{\bsnm{Tan}, \binits{M.}}:
\batitle{Coatnet: Marrying convolution and attention for all data sizes}.
\bjtitle{\textit{Proc. Neural Information Processing Systems (NeurIPS)}}
\bvolume{34},
\bfpage{3965}--\blpage{3977}
(\byear{2021})
\end{barticle}
\endbibitem

%%% 38
\bibitem[\protect\citeauthoryear{Jia et~al.}{2016}]{jia2016dynamic}
\begin{botherref}
\oauthor{\bsnm{Jia}, \binits{X.}},
\oauthor{\bsnm{De~Brabandere}, \binits{B.}},
\oauthor{\bsnm{Tuytelaars}, \binits{T.}},
\oauthor{\bsnm{Gool}, \binits{L.V.}}:
Dynamic filter networks.
\textit{Proc. Neural Information Processing Systems (NeurIPS)}
\textbf{29}
(2016)
\end{botherref}
\endbibitem

%%% 39
\bibitem[\protect\citeauthoryear{Li et~al.}{2021}]{li2021involution}
\begin{bchapter}
\bauthor{\bsnm{Li}, \binits{D.}},
\bauthor{\bsnm{Hu}, \binits{J.}},
\bauthor{\bsnm{Wang}, \binits{C.}},
\bauthor{\bsnm{Li}, \binits{X.}},
\bauthor{\bsnm{She}, \binits{Q.}},
\bauthor{\bsnm{Zhu}, \binits{L.}},
\bauthor{\bsnm{Zhang}, \binits{T.}},
\bauthor{\bsnm{Chen}, \binits{Q.}}:
\bctitle{Involution: Inverting the inherence of convolution for visual
  recognition}.
In: \bbtitle{\textit{Proc. IEEE Conference on Computer Vision and Pattern
  Recognition (CVPR)}},
pp. \bfpage{12321}--\blpage{12330}
(\byear{2021})
\end{bchapter}
\endbibitem

%%% 40
\bibitem[\protect\citeauthoryear{Chen et~al.}{2020}]{chen2020dynamic}
\begin{bchapter}
\bauthor{\bsnm{Chen}, \binits{Y.}},
\bauthor{\bsnm{Dai}, \binits{X.}},
\bauthor{\bsnm{Liu}, \binits{M.}},
\bauthor{\bsnm{Chen}, \binits{D.}},
\bauthor{\bsnm{Yuan}, \binits{L.}},
\bauthor{\bsnm{Liu}, \binits{Z.}}:
\bctitle{Dynamic convolution: Attention over convolution kernels}.
In: \bbtitle{\textit{Proc. IEEE Conference on Computer Vision and Pattern
  Recognition (CVPR)}},
pp. \bfpage{11030}--\blpage{11039}
(\byear{2020})
\end{bchapter}
\endbibitem

%%% 41
\bibitem[\protect\citeauthoryear{Ma et~al.}{2020}]{ma2020weightnet}
\begin{bchapter}
\bauthor{\bsnm{Ma}, \binits{N.}},
\bauthor{\bsnm{Zhang}, \binits{X.}},
\bauthor{\bsnm{Huang}, \binits{J.}},
\bauthor{\bsnm{Sun}, \binits{J.}}:
\bctitle{Weightnet: Revisiting the design space of weight networks}.
In: \bbtitle{\textit{Proc. European Conference on Computer Vision (ECCV)}},
pp. \bfpage{776}--\blpage{792}
(\byear{2020}).
\bcomment{Springer}
\end{bchapter}
\endbibitem

%%% 42
\bibitem[\protect\citeauthoryear{Shen et~al.}{2021}]{shen2021efficient}
\begin{bchapter}
\bauthor{\bsnm{Shen}, \binits{Z.}},
\bauthor{\bsnm{Zhang}, \binits{M.}},
\bauthor{\bsnm{Zhao}, \binits{H.}},
\bauthor{\bsnm{Yi}, \binits{S.}},
\bauthor{\bsnm{Li}, \binits{H.}}:
\bctitle{Efficient attention: Attention with linear complexities}.
In: \bbtitle{\textit{Proc. Winter Conference on Applications of Computer Vision
  (WACV)}},
pp. \bfpage{3531}--\blpage{3539}
(\byear{2021})
\end{bchapter}
\endbibitem

%%% 43
\bibitem[\protect\citeauthoryear{Katharopoulos
  et~al.}{2020}]{katharopoulos2020transformers}
\begin{bchapter}
\bauthor{\bsnm{Katharopoulos}, \binits{A.}},
\bauthor{\bsnm{Vyas}, \binits{A.}},
\bauthor{\bsnm{Pappas}, \binits{N.}},
\bauthor{\bsnm{Fleuret}, \binits{F.}}:
\bctitle{Transformers are rnns: Fast autoregressive transformers with linear
  attention}.
In: \bbtitle{\textit{Proc. International Conference on Machine Learning
  (ICML)}},
pp. \bfpage{5156}--\blpage{5165}
(\byear{2020}).
\bcomment{PMLR}
\end{bchapter}
\endbibitem

%%% 44
\bibitem[\protect\citeauthoryear{Rao et~al.}{2021}]{rao2021global}
\begin{barticle}
\bauthor{\bsnm{Rao}, \binits{Y.}},
\bauthor{\bsnm{Zhao}, \binits{W.}},
\bauthor{\bsnm{Zhu}, \binits{Z.}},
\bauthor{\bsnm{Lu}, \binits{J.}},
\bauthor{\bsnm{Zhou}, \binits{J.}}:
\batitle{Global filter networks for image classification}.
\bjtitle{\textit{Proc. Neural Information Processing Systems (NeurIPS)}}
\bvolume{34},
\bfpage{980}--\blpage{993}
(\byear{2021})
\end{barticle}
\endbibitem

%%% 45
\bibitem[\protect\citeauthoryear{Lee-Thorp et~al.}{2021}]{lee2021fnet}
\begin{botherref}
\oauthor{\bsnm{Lee-Thorp}, \binits{J.}},
\oauthor{\bsnm{Ainslie}, \binits{J.}},
\oauthor{\bsnm{Eckstein}, \binits{I.}},
\oauthor{\bsnm{Ontanon}, \binits{S.}}:
Fnet: Mixing tokens with fourier transforms.
\emph{arXiv preprint arXiv:2105.03824}
(2021)
\end{botherref}
\endbibitem

%%% 46
\bibitem[\protect\citeauthoryear{d’Ascoli et~al.}{2021}]{d2021convit}
\begin{bchapter}
\bauthor{\bsnm{d’Ascoli}, \binits{S.}},
\bauthor{\bsnm{Touvron}, \binits{H.}},
\bauthor{\bsnm{Leavitt}, \binits{M.L.}},
\bauthor{\bsnm{Morcos}, \binits{A.S.}},
\bauthor{\bsnm{Biroli}, \binits{G.}},
\bauthor{\bsnm{Sagun}, \binits{L.}}:
\bctitle{Convit: Improving vision transformers with soft convolutional
  inductive biases}.
In: \bbtitle{\textit{Proc. International Conference on Machine Learning
  (ICML)}},
pp. \bfpage{2286}--\blpage{2296}
(\byear{2021}).
\bcomment{PMLR}
\end{bchapter}
\endbibitem

%%% 47
\bibitem[\protect\citeauthoryear{Vaswani et~al.}{2017}]{vaswani2017attention}
\begin{botherref}
\oauthor{\bsnm{Vaswani}, \binits{A.}},
\oauthor{\bsnm{Shazeer}, \binits{N.}},
\oauthor{\bsnm{Parmar}, \binits{N.}},
\oauthor{\bsnm{Uszkoreit}, \binits{J.}},
\oauthor{\bsnm{Jones}, \binits{L.}},
\oauthor{\bsnm{Gomez}, \binits{A.N.}},
\oauthor{\bsnm{Kaiser}, \binits{{\L}.}},
\oauthor{\bsnm{Polosukhin}, \binits{I.}}:
Attention is all you need.
\textit{Proc. Neural Information Processing Systems (NeurIPS)}
\textbf{30}
(2017)
\end{botherref}
\endbibitem

%%% 48
\bibitem[\protect\citeauthoryear{Ramachandran
  et~al.}{2019}]{ramachandran2019stand}
\begin{botherref}
\oauthor{\bsnm{Ramachandran}, \binits{P.}},
\oauthor{\bsnm{Parmar}, \binits{N.}},
\oauthor{\bsnm{Vaswani}, \binits{A.}},
\oauthor{\bsnm{Bello}, \binits{I.}},
\oauthor{\bsnm{Levskaya}, \binits{A.}},
\oauthor{\bsnm{Shlens}, \binits{J.}}:
Stand-alone self-attention in vision models.
\textit{Proc. Neural Information Processing Systems (NeurIPS)}
\textbf{32}
(2019)
\end{botherref}
\endbibitem

%%% 49
\bibitem[\protect\citeauthoryear{Strang}{1986}]{strang1986proposal}
\begin{barticle}
\bauthor{\bsnm{Strang}, \binits{G.}}:
\batitle{A proposal for toeplitz matrix calculations}.
\bjtitle{\emph{Applied Mathematics}}
\bvolume{74}(\bissue{2}),
\bfpage{171}--\blpage{176}
(\byear{1986})
\end{barticle}
\endbibitem

%%% 50
\bibitem[\protect\citeauthoryear{Fu et~al.}{2022}]{fu2022hungry}
\begin{bchapter}
\bauthor{\bsnm{Fu}, \binits{D.Y.}},
\bauthor{\bsnm{Dao}, \binits{T.}},
\bauthor{\bsnm{Saab}, \binits{K.K.}},
\bauthor{\bsnm{Thomas}, \binits{A.W.}},
\bauthor{\bsnm{Rudra}, \binits{A.}},
\bauthor{\bsnm{Re}, \binits{C.}}:
\bctitle{Hungry hungry hippos: Towards language modeling with state space
  models}.
In: \bbtitle{\textit{Proc. International Conference on Learning Representations
  (ICLR)}}
(\byear{2022})
\end{bchapter}
\endbibitem

%%% 51
\bibitem[\protect\citeauthoryear{Russakovsky
  et~al.}{2015}]{imagenet15russakovsky}
\begin{barticle}
\bauthor{\bsnm{Russakovsky}, \binits{O.}},
\bauthor{\bsnm{Deng}, \binits{J.}},
\bauthor{\bsnm{Su}, \binits{H.}},
\bauthor{\bsnm{Krause}, \binits{J.}},
\bauthor{\bsnm{Satheesh}, \binits{S.}},
\bauthor{\bsnm{Ma}, \binits{S.}},
\bauthor{\bsnm{Huang}, \binits{Z.}},
\bauthor{\bsnm{Karpathy}, \binits{A.}},
\bauthor{\bsnm{Khosla}, \binits{A.}},
\bauthor{\bsnm{Bernstein}, \binits{M.}},
\bauthor{\bsnm{Berg}, \binits{A.C.}},
\bauthor{\bsnm{Fei-Fei}, \binits{L.}}:
\batitle{{ImageNet Large Scale Visual Recognition Challenge}}.
\bjtitle{\textit{International Journal of Computer Vision (IJCV)}}
\bvolume{115}(\bissue{3}),
\bfpage{211}--\blpage{252}
(\byear{2015})
\doiurl{10.1007/s11263-015-0816-y}
\end{barticle}
\endbibitem

%%% 52
\bibitem[\protect\citeauthoryear{Lin et~al.}{2023}]{lin2023scale}
\begin{bchapter}
\bauthor{\bsnm{Lin}, \binits{W.}},
\bauthor{\bsnm{Wu}, \binits{Z.}},
\bauthor{\bsnm{Chen}, \binits{J.}},
\bauthor{\bsnm{Huang}, \binits{J.}},
\bauthor{\bsnm{Jin}, \binits{L.}}:
\bctitle{Scale-aware modulation meet transformer}.
In: \bbtitle{\textit{Proc. IEEE International Conference on Computer Vision
  (ICCV)}},
pp. \bfpage{6015}--\blpage{6026}
(\byear{2023})
\end{bchapter}
\endbibitem

%%% 53
\bibitem[\protect\citeauthoryear{Li et~al.}{2023}]{li2023uniformer}
\begin{barticle}
\bauthor{\bsnm{Li}, \binits{K.}},
\bauthor{\bsnm{Wang}, \binits{Y.}},
\bauthor{\bsnm{Zhang}, \binits{J.}},
\bauthor{\bsnm{Gao}, \binits{P.}},
\bauthor{\bsnm{Song}, \binits{G.}},
\bauthor{\bsnm{Liu}, \binits{Y.}},
\bauthor{\bsnm{Li}, \binits{H.}},
\bauthor{\bsnm{Qiao}, \binits{Y.}}:
\batitle{Uniformer: Unifying convolution and self-attention for visual
  recognition}.
\bjtitle{\textit{IEEE Transactions on Pattern Analysis and Machine Intelligence
  (TPAMI)}}
\bvolume{45}(\bissue{10}),
\bfpage{12581}--\blpage{12600}
(\byear{2023})
\end{barticle}
\endbibitem

%%% 54
\bibitem[\protect\citeauthoryear{Wang et~al.}{2023}]{wang2023internimage}
\begin{bchapter}
\bauthor{\bsnm{Wang}, \binits{W.}},
\bauthor{\bsnm{Dai}, \binits{J.}},
\bauthor{\bsnm{Chen}, \binits{Z.}},
\bauthor{\bsnm{Huang}, \binits{Z.}},
\bauthor{\bsnm{Li}, \binits{Z.}},
\bauthor{\bsnm{Zhu}, \binits{X.}},
\bauthor{\bsnm{Hu}, \binits{X.}},
\bauthor{\bsnm{Lu}, \binits{T.}},
\bauthor{\bsnm{Lu}, \binits{L.}},
\bauthor{\bsnm{Li}, \binits{H.}}, \betal:
\bctitle{Internimage: Exploring large-scale vision foundation models with
  deformable convolutions}.
In: \bbtitle{\textit{Proc. IEEE Conference on Computer Vision and Pattern
  Recognition (CVPR)}},
pp. \bfpage{14408}--\blpage{14419}
(\byear{2023})
\end{bchapter}
\endbibitem

%%% 55
\bibitem[\protect\citeauthoryear{Pan et~al.}{2023}]{pan2023slide}
\begin{bchapter}
\bauthor{\bsnm{Pan}, \binits{X.}},
\bauthor{\bsnm{Ye}, \binits{T.}},
\bauthor{\bsnm{Xia}, \binits{Z.}},
\bauthor{\bsnm{Song}, \binits{S.}},
\bauthor{\bsnm{Huang}, \binits{G.}}:
\bctitle{Slide-transformer: Hierarchical vision transformer with local
  self-attention}.
In: \bbtitle{\textit{Proc. IEEE Conference on Computer Vision and Pattern
  Recognition (CVPR)}},
pp. \bfpage{2082}--\blpage{2091}
(\byear{2023})
\end{bchapter}
\endbibitem

%%% 56
\bibitem[\protect\citeauthoryear{Chu et~al.}{2023}]{chuconditional}
\begin{bchapter}
\bauthor{\bsnm{Chu}, \binits{X.}},
\bauthor{\bsnm{Tian}, \binits{Z.}},
\bauthor{\bsnm{Zhang}, \binits{B.}},
\bauthor{\bsnm{Wang}, \binits{X.}},
\bauthor{\bsnm{Shen}, \binits{C.}}:
\bctitle{Conditional positional encodings for vision transformers}.
In: \bbtitle{\textit{Proc. International Conference on Learning Representations
  (ICLR)}}
(\byear{2023})
\end{bchapter}
\endbibitem

%%% 57
\bibitem[\protect\citeauthoryear{Howard et~al.}{2017}]{howard2017mobilenets}
\begin{botherref}
\oauthor{\bsnm{Howard}, \binits{A.G.}},
\oauthor{\bsnm{Zhu}, \binits{M.}},
\oauthor{\bsnm{Chen}, \binits{B.}},
\oauthor{\bsnm{Kalenichenko}, \binits{D.}},
\oauthor{\bsnm{Wang}, \binits{W.}},
\oauthor{\bsnm{Weyand}, \binits{T.}},
\oauthor{\bsnm{Andreetto}, \binits{M.}},
\oauthor{\bsnm{Adam}, \binits{H.}}:
Mobilenets: Efficient convolutional neural networks for mobile vision
  applications.
\emph{arXiv preprint arXiv:1704.04861}
(2017)
\end{botherref}
\endbibitem

%%% 58
\bibitem[\protect\citeauthoryear{He et~al.}{2016}]{he2016deep}
\begin{bchapter}
\bauthor{\bsnm{He}, \binits{K.}},
\bauthor{\bsnm{Zhang}, \binits{X.}},
\bauthor{\bsnm{Ren}, \binits{S.}},
\bauthor{\bsnm{Sun}, \binits{J.}}:
\bctitle{Deep residual learning for image recognition}.
In: \bbtitle{\textit{Proc. IEEE Conference on Computer Vision and Pattern
  Recognition (CVPR)}},
pp. \bfpage{770}--\blpage{778}
(\byear{2016})
\end{bchapter}
\endbibitem

%%% 59
\bibitem[\protect\citeauthoryear{Radosavovic
  et~al.}{2020}]{radosavovic2020designing}
\begin{bchapter}
\bauthor{\bsnm{Radosavovic}, \binits{I.}},
\bauthor{\bsnm{Kosaraju}, \binits{R.P.}},
\bauthor{\bsnm{Girshick}, \binits{R.}},
\bauthor{\bsnm{He}, \binits{K.}},
\bauthor{\bsnm{Doll{\'a}r}, \binits{P.}}:
\bctitle{Designing network design spaces}.
In: \bbtitle{\textit{Proc. IEEE Conference on Computer Vision and Pattern
  Recognition (CVPR)}},
pp. \bfpage{10428}--\blpage{10436}
(\byear{2020})
\end{bchapter}
\endbibitem

%%% 60
\bibitem[\protect\citeauthoryear{Liu et~al.}{2022}]{liu2022convnet}
\begin{bchapter}
\bauthor{\bsnm{Liu}, \binits{Z.}},
\bauthor{\bsnm{Mao}, \binits{H.}},
\bauthor{\bsnm{Wu}, \binits{C.-Y.}},
\bauthor{\bsnm{Feichtenhofer}, \binits{C.}},
\bauthor{\bsnm{Darrell}, \binits{T.}},
\bauthor{\bsnm{Xie}, \binits{S.}}:
\bctitle{A convnet for the 2020s}.
In: \bbtitle{\textit{Proc. IEEE Conference on Computer Vision and Pattern
  Recognition (CVPR)}},
pp. \bfpage{11976}--\blpage{11986}
(\byear{2022})
\end{bchapter}
\endbibitem

%%% 61
\bibitem[\protect\citeauthoryear{Wang et~al.}{2021}]{wang2021pyramid}
\begin{bchapter}
\bauthor{\bsnm{Wang}, \binits{W.}},
\bauthor{\bsnm{Xie}, \binits{E.}},
\bauthor{\bsnm{Li}, \binits{X.}},
\bauthor{\bsnm{Fan}, \binits{D.-P.}},
\bauthor{\bsnm{Song}, \binits{K.}},
\bauthor{\bsnm{Liang}, \binits{D.}},
\bauthor{\bsnm{Lu}, \binits{T.}},
\bauthor{\bsnm{Luo}, \binits{P.}},
\bauthor{\bsnm{Shao}, \binits{L.}}:
\bctitle{Pyramid vision transformer: A versatile backbone for dense prediction
  without convolutions}.
In: \bbtitle{\textit{Proc. IEEE International Conference on Computer Vision
  (ICCV)}},
pp. \bfpage{568}--\blpage{578}
(\byear{2021})
\end{bchapter}
\endbibitem

%%% 62
\bibitem[\protect\citeauthoryear{Guo et~al.}{2022}]{guo2022visual}
\begin{botherref}
\oauthor{\bsnm{Guo}, \binits{M.-H.}},
\oauthor{\bsnm{Lu}, \binits{C.-Z.}},
\oauthor{\bsnm{Liu}, \binits{Z.-N.}},
\oauthor{\bsnm{Cheng}, \binits{M.-M.}},
\oauthor{\bsnm{Hu}, \binits{S.-M.}}:
Visual attention network.
\emph{arXiv preprint arXiv:2202.09741}
(2022)
\end{botherref}
\endbibitem

%%% 63
\bibitem[\protect\citeauthoryear{Vasu et~al.}{2023}]{vasu2023fastvit}
\begin{bchapter}
\bauthor{\bsnm{Vasu}, \binits{P.K.A.}},
\bauthor{\bsnm{Gabriel}, \binits{J.}},
\bauthor{\bsnm{Zhu}, \binits{J.}},
\bauthor{\bsnm{Tuzel}, \binits{O.}},
\bauthor{\bsnm{Ranjan}, \binits{A.}}:
\bctitle{Fastvit: A fast hybrid vision transformer using structural
  reparameterization}.
In: \bbtitle{\textit{Proc. IEEE International Conference on Computer Vision
  (ICCV)}},
pp. \bfpage{5785}--\blpage{5795}
(\byear{2023})
\end{bchapter}
\endbibitem

%%% 64
\bibitem[\protect\citeauthoryear{Feichtenhofer}{2020}]{feichtenhofer2020x3d}
\begin{bchapter}
\bauthor{\bsnm{Feichtenhofer}, \binits{C.}}:
\bctitle{X3d: Expanding architectures for efficient video recognition}.
In: \bbtitle{\textit{Proc. IEEE Conference on Computer Vision and Pattern
  Recognition (CVPR)}}
(\byear{2020})
\end{bchapter}
\endbibitem

%%% 65
\bibitem[\protect\citeauthoryear{Feichtenhofer
  et~al.}{2019}]{feichtenhofer2019slowfast}
\begin{bchapter}
\bauthor{\bsnm{Feichtenhofer}, \binits{C.}},
\bauthor{\bsnm{Fan}, \binits{H.}},
\bauthor{\bsnm{Malik}, \binits{J.}},
\bauthor{\bsnm{He}, \binits{K.}}:
\bctitle{Slowfast networks for video recognition}.
In: \bbtitle{\textit{Proc. IEEE International Conference on Computer Vision
  (ICCV)}}
(\byear{2019})
\end{bchapter}
\endbibitem

%%% 66
\bibitem[\protect\citeauthoryear{Bulat et~al.}{2021}]{bulat2021space}
\begin{barticle}
\bauthor{\bsnm{Bulat}, \binits{A.}},
\bauthor{\bsnm{Perez~Rua}, \binits{J.M.}},
\bauthor{\bsnm{Sudhakaran}, \binits{S.}},
\bauthor{\bsnm{Martinez}, \binits{B.}},
\bauthor{\bsnm{Tzimiropoulos}, \binits{G.}}:
\batitle{Space-time mixing attention for video transformer}.
\bjtitle{\textit{Proc. Neural Information Processing Systems (NeurIPS)}}
\bvolume{34},
\bfpage{19594}--\blpage{19607}
(\byear{2021})
\end{barticle}
\endbibitem

%%% 67
\bibitem[\protect\citeauthoryear{Patrick et~al.}{2021}]{patrick2021keeping}
\begin{barticle}
\bauthor{\bsnm{Patrick}, \binits{M.}},
\bauthor{\bsnm{Campbell}, \binits{D.}},
\bauthor{\bsnm{Asano}, \binits{Y.}},
\bauthor{\bsnm{Misra}, \binits{I.}},
\bauthor{\bsnm{Metze}, \binits{F.}},
\bauthor{\bsnm{Feichtenhofer}, \binits{C.}},
\bauthor{\bsnm{Vedaldi}, \binits{A.}},
\bauthor{\bsnm{Henriques}, \binits{J.F.}}:
\batitle{Keeping your eye on the ball: Trajectory attention in video
  transformers}.
\bjtitle{\textit{Proc. Neural Information Processing Systems (NeurIPS)}}
\bvolume{34},
\bfpage{12493}--\blpage{12506}
(\byear{2021})
\end{barticle}
\endbibitem

%%% 68
\bibitem[\protect\citeauthoryear{Liu et~al.}{2022}]{liu2022video}
\begin{bchapter}
\bauthor{\bsnm{Liu}, \binits{Z.}},
\bauthor{\bsnm{Ning}, \binits{J.}},
\bauthor{\bsnm{Cao}, \binits{Y.}},
\bauthor{\bsnm{Wei}, \binits{Y.}},
\bauthor{\bsnm{Zhang}, \binits{Z.}},
\bauthor{\bsnm{Lin}, \binits{S.}},
\bauthor{\bsnm{Hu}, \binits{H.}}:
\bctitle{Video swin transformer}.
In: \bbtitle{\emph{Proceedings of the IEEE/CVF Conference on Computer Vision
  and Pattern Recognition}},
pp. \bfpage{3202}--\blpage{3211}
(\byear{2022})
\end{bchapter}
\endbibitem

%%% 69
\bibitem[\protect\citeauthoryear{Bertasius et~al.}{2021}]{bertasius2021space}
\begin{bchapter}
\bauthor{\bsnm{Bertasius}, \binits{G.}},
\bauthor{\bsnm{Wang}, \binits{H.}},
\bauthor{\bsnm{Torresani}, \binits{L.}}:
\bctitle{Is space-time attention all you need for video understanding?}
In: \bbtitle{\textit{Proc. International Conference on Machine Learning
  (ICML)}},
vol. \bseriesno{2},
p. \bfpage{4}
(\byear{2021})
\end{bchapter}
\endbibitem

%%% 70
\bibitem[\protect\citeauthoryear{Fan et~al.}{2021}]{fan2021multiscale}
\begin{bchapter}
\bauthor{\bsnm{Fan}, \binits{H.}},
\bauthor{\bsnm{Xiong}, \binits{B.}},
\bauthor{\bsnm{Mangalam}, \binits{K.}},
\bauthor{\bsnm{Li}, \binits{Y.}},
\bauthor{\bsnm{Yan}, \binits{Z.}},
\bauthor{\bsnm{Malik}, \binits{J.}},
\bauthor{\bsnm{Feichtenhofer}, \binits{C.}}:
\bctitle{Multiscale vision transformers}.
In: \bbtitle{\textit{Proc. IEEE International Conference on Computer Vision
  (ICCV)}},
pp. \bfpage{6824}--\blpage{6835}
(\byear{2021})
\end{bchapter}
\endbibitem

%%% 71
\bibitem[\protect\citeauthoryear{Chu et~al.}{2021}]{chu2021twins}
\begin{barticle}
\bauthor{\bsnm{Chu}, \binits{X.}},
\bauthor{\bsnm{Tian}, \binits{Z.}},
\bauthor{\bsnm{Wang}, \binits{Y.}},
\bauthor{\bsnm{Zhang}, \binits{B.}},
\bauthor{\bsnm{Ren}, \binits{H.}},
\bauthor{\bsnm{Wei}, \binits{X.}},
\bauthor{\bsnm{Xia}, \binits{H.}},
\bauthor{\bsnm{Shen}, \binits{C.}}:
\batitle{Twins: Revisiting the design of spatial attention in vision
  transformers}.
\bjtitle{\textit{Proc. Neural Information Processing Systems (NeurIPS)}}
\bvolume{34},
\bfpage{9355}--\blpage{9366}
(\byear{2021})
\end{barticle}
\endbibitem

%%% 72
\bibitem[\protect\citeauthoryear{Zhang et~al.}{2021}]{zhang2021multi}
\begin{bchapter}
\bauthor{\bsnm{Zhang}, \binits{P.}},
\bauthor{\bsnm{Dai}, \binits{X.}},
\bauthor{\bsnm{Yang}, \binits{J.}},
\bauthor{\bsnm{Xiao}, \binits{B.}},
\bauthor{\bsnm{Yuan}, \binits{L.}},
\bauthor{\bsnm{Zhang}, \binits{L.}},
\bauthor{\bsnm{Gao}, \binits{J.}}:
\bctitle{Multi-scale vision longformer: A new vision transformer for
  high-resolution image encoding}.
In: \bbtitle{\textit{Proc. IEEE International Conference on Computer Vision
  (ICCV)}},
pp. \bfpage{2998}--\blpage{3008}
(\byear{2021})
\end{bchapter}
\endbibitem

%%% 73
\bibitem[\protect\citeauthoryear{Yang et~al.}{2021}]{yang2021focal}
\begin{barticle}
\bauthor{\bsnm{Yang}, \binits{J.}},
\bauthor{\bsnm{Li}, \binits{C.}},
\bauthor{\bsnm{Zhang}, \binits{P.}},
\bauthor{\bsnm{Dai}, \binits{X.}},
\bauthor{\bsnm{Xiao}, \binits{B.}},
\bauthor{\bsnm{Yuan}, \binits{L.}},
\bauthor{\bsnm{Gao}, \binits{J.}}:
\batitle{Focal attention for long-range interactions in vision transformers}.
\bjtitle{\textit{Proc. Neural Information Processing Systems (NeurIPS)}}
\bvolume{34},
\bfpage{30008}--\blpage{30022}
(\byear{2021})
\end{barticle}
\endbibitem

%%% 74
\bibitem[\protect\citeauthoryear{He et~al.}{2017}]{he2017mask}
\begin{bchapter}
\bauthor{\bsnm{He}, \binits{K.}},
\bauthor{\bsnm{Gkioxari}, \binits{G.}},
\bauthor{\bsnm{Doll{\'a}r}, \binits{P.}},
\bauthor{\bsnm{Girshick}, \binits{R.}}:
\bctitle{Mask r-cnn}.
In: \bbtitle{\textit{Proc. IEEE International Conference on Computer Vision
  (ICCV)}},
pp. \bfpage{2961}--\blpage{2969}
(\byear{2017})
\end{bchapter}
\endbibitem

%%% 75
\bibitem[\protect\citeauthoryear{Kirillov et~al.}{2019}]{kirillov2019panoptic}
\begin{bchapter}
\bauthor{\bsnm{Kirillov}, \binits{A.}},
\bauthor{\bsnm{Girshick}, \binits{R.}},
\bauthor{\bsnm{He}, \binits{K.}},
\bauthor{\bsnm{Doll{\'a}r}, \binits{P.}}:
\bctitle{Panoptic feature pyramid networks}.
In: \bbtitle{\textit{Proc. IEEE Conference on Computer Vision and Pattern
  Recognition (CVPR)}},
pp. \bfpage{6399}--\blpage{6408}
(\byear{2019})
\end{bchapter}
\endbibitem

%%% 76
\bibitem[\protect\citeauthoryear{Loshchilov and
  Hutter}{2018}]{loshchilov2018decoupled}
\begin{bchapter}
\bauthor{\bsnm{Loshchilov}, \binits{I.}},
\bauthor{\bsnm{Hutter}, \binits{F.}}:
\bctitle{Decoupled weight decay regularization}.
In: \bbtitle{\textit{Proc. International Conference on Learning Representations
  (ICLR)}}
(\byear{2018})
\end{bchapter}
\endbibitem

%%% 77
\bibitem[\protect\citeauthoryear{Zhang et~al.}{2018}]{zhang2018mixup}
\begin{bchapter}
\bauthor{\bsnm{Zhang}, \binits{H.}},
\bauthor{\bsnm{Cisse}, \binits{M.}},
\bauthor{\bsnm{Dauphin}, \binits{Y.N.}},
\bauthor{\bsnm{Lopez-Paz}, \binits{D.}}:
\bctitle{mixup: Beyond empirical risk minimization}.
In: \bbtitle{\textit{Proc. International Conference on Learning Representations
  (ICLR)}}
(\byear{2018})
\end{bchapter}
\endbibitem

%%% 78
\bibitem[\protect\citeauthoryear{Yun et~al.}{2019}]{yun2019cutmix}
\begin{bchapter}
\bauthor{\bsnm{Yun}, \binits{S.}},
\bauthor{\bsnm{Han}, \binits{D.}},
\bauthor{\bsnm{Oh}, \binits{S.J.}},
\bauthor{\bsnm{Chun}, \binits{S.}},
\bauthor{\bsnm{Choe}, \binits{J.}},
\bauthor{\bsnm{Yoo}, \binits{Y.}}:
\bctitle{Cutmix: Regularization strategy to train strong classifiers with
  localizable features}.
In: \bbtitle{\textit{Proc. IEEE International Conference on Computer Vision
  (ICCV)}},
pp. \bfpage{6023}--\blpage{6032}
(\byear{2019})
\end{bchapter}
\endbibitem

%%% 79
\bibitem[\protect\citeauthoryear{Szegedy et~al.}{2016}]{szegedy2016rethinking}
\begin{bchapter}
\bauthor{\bsnm{Szegedy}, \binits{C.}},
\bauthor{\bsnm{Vanhoucke}, \binits{V.}},
\bauthor{\bsnm{Ioffe}, \binits{S.}},
\bauthor{\bsnm{Shlens}, \binits{J.}},
\bauthor{\bsnm{Wojna}, \binits{Z.}}:
\bctitle{Rethinking the inception architecture for computer vision}.
In: \bbtitle{\textit{Proc. IEEE Conference on Computer Vision and Pattern
  Recognition (CVPR)}},
pp. \bfpage{2818}--\blpage{2826}
(\byear{2016})
\end{bchapter}
\endbibitem

%%% 80
\bibitem[\protect\citeauthoryear{Huang et~al.}{2016}]{huang2016deep}
\begin{bchapter}
\bauthor{\bsnm{Huang}, \binits{G.}},
\bauthor{\bsnm{Sun}, \binits{Y.}},
\bauthor{\bsnm{Liu}, \binits{Z.}},
\bauthor{\bsnm{Sedra}, \binits{D.}},
\bauthor{\bsnm{Weinberger}, \binits{K.Q.}}:
\bctitle{Deep networks with stochastic depth}.
In: \bbtitle{\textit{Proc. European Conference on Computer Vision (ECCV)}},
pp. \bfpage{646}--\blpage{661}
(\byear{2016}).
\bcomment{Springer}
\end{bchapter}
\endbibitem

%%% 81
\bibitem[\protect\citeauthoryear{Cubuk et~al.}{2020}]{cubuk2020randaugment}
\begin{barticle}
\bauthor{\bsnm{Cubuk}, \binits{E.D.}},
\bauthor{\bsnm{Zoph}, \binits{B.}},
\bauthor{\bsnm{Shlens}, \binits{J.}},
\bauthor{\bsnm{Le}, \binits{Q.}}:
\batitle{Randaugment: Practical automated data augmentation with a reduced
  search space}.
\bjtitle{\textit{Proc. Neural Information Processing Systems (NeurIPS)}}
\bvolume{33},
\bfpage{18613}--\blpage{18624}
(\byear{2020})
\end{barticle}
\endbibitem

%%% 82
\bibitem[\protect\citeauthoryear{Zhong et~al.}{2020}]{zhong2020random}
\begin{bchapter}
\bauthor{\bsnm{Zhong}, \binits{Z.}},
\bauthor{\bsnm{Zheng}, \binits{L.}},
\bauthor{\bsnm{Kang}, \binits{G.}},
\bauthor{\bsnm{Li}, \binits{S.}},
\bauthor{\bsnm{Yang}, \binits{Y.}}:
\bctitle{Random erasing data augmentation}.
In: \bbtitle{\textit{Proc. AAAI Conference on Artificial Intelligence (AAAI)}},
vol. \bseriesno{34},
pp. \bfpage{13001}--\blpage{13008}
(\byear{2020})
\end{bchapter}
\endbibitem

%%% 83
\bibitem[\protect\citeauthoryear{Wang et~al.}{2018}]{wang2018non}
\begin{bchapter}
\bauthor{\bsnm{Wang}, \binits{X.}},
\bauthor{\bsnm{Girshick}, \binits{R.}},
\bauthor{\bsnm{Gupta}, \binits{A.}},
\bauthor{\bsnm{He}, \binits{K.}}:
\bctitle{Non-local neural networks}.
In: \bbtitle{Proceedings of the IEEE Conference on Computer Vision and Pattern
  Recognition},
pp. \bfpage{7794}--\blpage{7803}
(\byear{2018})
\end{bchapter}
\endbibitem

%%% 84
\bibitem[\protect\citeauthoryear{Wang et~al.}{2016}]{wang2016temporal}
\begin{bchapter}
\bauthor{\bsnm{Wang}, \binits{L.}},
\bauthor{\bsnm{Xiong}, \binits{Y.}},
\bauthor{\bsnm{Wang}, \binits{Z.}},
\bauthor{\bsnm{Qiao}, \binits{Y.}},
\bauthor{\bsnm{Lin}, \binits{D.}},
\bauthor{\bsnm{Tang}, \binits{X.}},
\bauthor{\bsnm{Van~Gool}, \binits{L.}}:
\bctitle{Temporal segment networks: Towards good practices for deep action
  recognition}.
In: \bbtitle{\textit{Proc. European Conference on Computer Vision (ECCV)}},
pp. \bfpage{20}--\blpage{36}
(\byear{2016}).
\bcomment{Springer}
\end{bchapter}
\endbibitem

%%% 85
\bibitem[\protect\citeauthoryear{Fu et~al.}{2023}]{fu2023flashfftconv}
\begin{botherref}
\oauthor{\bsnm{Fu}, \binits{D.Y.}},
\oauthor{\bsnm{Kumbong}, \binits{H.}},
\oauthor{\bsnm{Nguyen}, \binits{E.}},
\oauthor{\bsnm{R{\'e}}, \binits{C.}}:
Flashfftconv: Efficient convolutions for long sequences with tensor cores.
arXiv preprint arXiv:2311.05908
(2023)
\end{botherref}
\endbibitem

\end{thebibliography}
\clearpage
\appendix

% !TEX root = ../main.tex
\onecolumn
\begin{center}
\textbf{\large Supplementary Material of "Lightweight Structure-Aware Attention for Visual Understanding"}
\end{center}

\section{Implementation details}
%\textbf{LiSA details.}

\subsection{Architecture Details}

\textbf{LiSA.}
In addition to the details included in the main paper, we provide pseudo-code of structure-aware attention (Eq.~4 \& Eq.~5 in the main paper) and LiSA in Fig.~\ref{fig:SA_pseudo} and Fig.~\ref{fig:LiSA_pseudo}, respectively.
The notation of multi-head is omitted for clarity and simplicity.
As shown in the pseudo-code, we effectively reduce the computational complexity with the FFT approximation.
% In actual implementation, we employ 2D RFFT/IRFFT for images and 3D variant for videos.

\begin{figure}[t]
\centering
    \includegraphics[width=0.45\columnwidth]{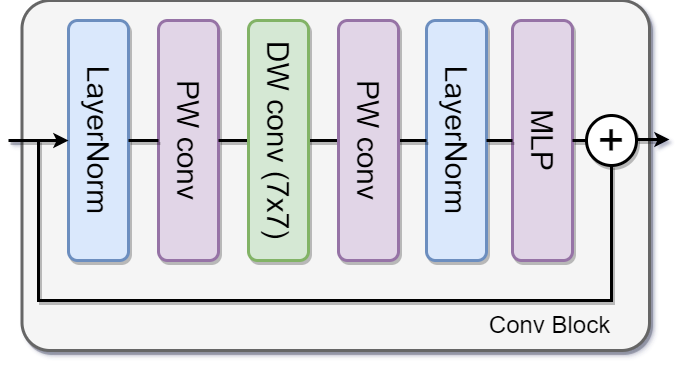}
    \caption{\textbf{Convolutional block in hybrid models.} `PW conv' denotes a pointwise convolution and `DW conv' denotes a depthwise convolution.}
    \label{fig:convblock}
\end{figure}

\textbf{LiSANet.}
A detailed overview of our proposed architectures is shown in Tab.~\ref{table:arch2}. LiSANet-I is composed of a single stage following the traditional ViT guidelines~\citep{arnab2021vivit,touvron2021training}. The number of tokens is constant in this model: $14\times14$. Focusing on the blocks, we have an initial patch embedding layer with a stride of 16 pixels and then, 12 LiSA blocks with an embedding size of 192.
On the other hand, our hierarchical ViTs follow the guidelines proposed in~\citep{liu2021swin,rao2021global,dong2022cswin}.
Both models (LiSANet-S and LiSANet-B) are composed of the same four stages with different token sizes and numbers of tokens.
For hybrid models (HyLiSANet-S and HyLiSANet-B), we follow the strategies proposed in~\citep{li2022uniformer,dai2021coatnet}.
We adopt convolutional blocks for the early two stages and employ LiSA blocks for the other stages.
Each convolutional block consists of two pointwise convolutions and one depthwise convolution as illustrated in Fig.~\ref{fig:convblock}.
We adopt the overlapping patch embedding strategy~\citep{wang2022pvt} for hierarchical models.
MLP ratios are set to 4 for the early two stages and 3 for the last two stages.
We adopt convolutional position embedding~\citep{chuconditional} and convolutional MLP~\citep{wang2022pvt} for our hierarchical models.
% The number of LiSA blocks differs in both models, with LiSANet-B the larger model employing a larger number of blocks (16 vs.\ 24 blocks).

\subsection{Experimental Setup}
\textbf{Image classification (ImageNet-1K).}
% Our models are trained using Pytorch 1.11~\cite{paszke2019pytorch}.
Our models are trained with AdamW~\citep{loshchilov2018decoupled} with a weight decay of 0.05 and a learning rate of $\frac{0.0005}{512} \cdot batch\_size$ with a cosine decay scheduler and 20 warm-up epochs.
Our isotropic model (LiSANet-I) is trained for 150 epochs and hierarchical models are trained for 300 epochs. Following the training recipe proposed in~\citep{liu2021swin}, we apply several regularization techniques such as Mixup~\citep{zhang2018mixup}, Cutmix~\citep{yun2019cutmix}, label smoothing~\citep{szegedy2016rethinking} and stochastic depth~\citep{huang2016deep}.
The stochastic depth strategy is applied only for the hierarchical models with a probability of 0.1 and 0.4 for the LiSANet-S and LiSANet-B models, respectively. In addition, we also apply several data augmentation techniques like Rand-Augment~\citep{cubuk2020randaugment}, random erasing~\citep{zhong2020random}, and repeated augmentation.
Note that all these hyperparameter values and data-augmentation techniques are selected following the training recipes of the previous works~\citep{liu2021swin,touvron2021training}.

\textbf{Image classification (ImageNet-21K).}
For ImageNet-21K pre-training, models are trained with AdamW~\citep{loshchilov2018decoupled} with a weight decay of 0.05 and a base learning rate of 4e-3 \& batch size 4096 with a cosine decay scheduler and 5 warm-up epochs. The total number of epochs are set to 90, and the stochastic depth rate is set to 0.1 \& 0.2 for HyLiSANet-S \& -B models, respectively. For the other details, we follow the training setup proposed in~\citep{liu2022convnet}.
For ImageNet-1K fine-tuning, models are trained with AdamW~\citep{loshchilov2018decoupled} with a weight decay of 1e-8 and a base learning rate of 5e-5 \& batch size 512 with a cosine decay scheduler. The total number of epochs are set to 30, and the stochastic depth rate is set to 0.1 \& 0.2 for HyLiSANet-S \& -B models, respectively. For the other details, we follow the training setup proposed in~\citep{liu2022convnet}.

\textbf{Video action recognition (Kinetics-400).}
ImageNet-1K pre-trained models (HyLiSANet-S, HyLiSANet-B) are adapted to video models and utilized as backbones, and the models are trained with  AdamW~\citep{loshchilov2018decoupled} using a weight decay of 0.05 and $\frac{0.0001}{32} \cdot batch\_size$ with a cosine decay scheduler and 10 warm-up epochs. The total number of epochs are set to 110, and the stochastic depth rate is set to 0.15 \& 0.3 for HyLiSANet-S \& -B models, respectively.
For the other regularization or data-augmentation details, we follow the training setup proposed in~\citep{li2023uniformer}.
Number of input frames is set to 16, and we adopt the dense sampling strategy~\citep{wang2018non} for training and multi-clip (16$\times$1$\times$4) inference for testing. All scores are averaged for the final result.

\textbf{Video action recognition (Something-V2).}
Video models are trained with AdamW~\citep{loshchilov2018decoupled} using a weight decay of 0.05 and $\frac{0.0002}{32} \cdot batch\_size$ with a cosine decay scheduler and 5 warm-up epochs.
The total number of epochs is set to 60, and we apply several regularization or data-augmentation techniques following~\citep{li2022mvitv2,li2022uniformer,arnab2021vivit}.
We adopt the uniform sampling strategy~\citep{wang2016temporal} for training and a single crop inference for testing.

\textbf{Object detection (COCO).}
We adopt standard Mask R-CNN~\citep{he2017mask} detection frameworks, and ImageNet-1K pre-trained models (LiSANet-S, HyLiSANet-S) are utilized as backbones.
% weights for fine-tuning.
We use a 1$\times$ schedule (12 epochs) with total batch size 16, and follow the same recipe as in~\citep{liu2021swin}.
For training, the shorter side of the image is resized to 800 pixels while keeping the longer side no more than 1333 pixels.
AdamW~\citep{loshchilov2018decoupled} with a weight decay of 0.05 is adopted as an optimizer, and the initial learning rate is set to 0.0001.
The stochastic depth rate is set to 0.1 and we follow the other details proposed in~\citep{liu2021swin}.

\textbf{Semantic segmentation (ADE-20K).}
We adopt Semantic FPN~\citep{kirillov2019panoptic} as a basic framework.
ImageNet-1K pre-trained models (LiSANet-S, HyLiSANet-S) are utilized as backbones.
The framework is trained for 80k iterations with a cosine decay scheduler.
The stochastic depth rate is set to 0.15 and we follow the other details proposed in~\citep{wang2022pvt} for a fair comparison.

\begin{figure}[t]
% \small
\begin{lstlisting}[language=python]
# B: batches, N: tokens, C: channels, D: latent_channels
def structure_aware_attn_Eq4(input, e): 
# input shape: [B,N,C], e shape: [2N-1]
    qkv = linear_proj(input, channels=3C) # shape: [B,N,3C]
    query,key,value = split(qkv, [C,C,C], dim=-1)
    # query,key,value shape: [B,N,C]
    query = L2norm(query) # shape: [B,N,C]
    key = L2norm(key) # shape: [B,N,C]
    
    R = Toeplitz(e) # shape: [N,N]
    attn = einsum(query,key, 'BNC,BMC->BNM') # shape: [B,N,N]
    attn_R = attn * R # shape: [B,N,N]
    
    out = einsum(attn_R,value,'BNM,BMC->BNC') # shape: [B,N,C]
    out = linear_proj2(out, channels=C) # shape: [B,N,C]
    return out    
    
def structure_aware_attn_Eq5(input, Ea, Eb, B): 
# input shape: [B,N,C], Ea,Eb shape: [2N-1,D], Bb shape: [C,D]
    qkv = linear_proj(input, channels=3C) # shape: [B,N,3C]
    query,key,value = split(qkv, [C,C,C], dim=-1)
    # query,key,value shape: [B,N,C]
    query = L2norm(query) # shape: [B,N,C]
    key = L2norm(key) # shape: [B,N,C]
    
    Ra = Toeplitz(Ea) # shape: [N,N,D]
    Rb = Toeplitz(Eb) # shape: [N,N,D]
    K_Ra = einsum(key,Ra,'BMC,NMD->BNCD') # shape: [B,N,C,D]
    Rb_V = einsum(Rb,value, 'NMD,BMV->BNVD') # shape: [B,N,C,D]
    Rb_V_B = Rb_V + B # shape: [B,N,C,D]
    
    out = einsum(query,K_Ra,Rb_v_B,'BNC,BNCD,BNVD->BNV') # shape: [B,N,C]
    out = linear_proj2(out, channels=C) # shape: [B,N,C]
    return out    
\end{lstlisting}
    \caption{
    \textbf{Pseudo-code for structure-aware attention.} We describe the way of learning convolutional inductive biases in structure-aware attention (Eq.~4) and its basic form (Eq.~5) presented in Sec.~4.1 of the main paper.
    }
    \label{fig:SA_pseudo}
\end{figure}

\begin{figure}[t]
\small
\begin{lstlisting}[language=python]
# B: batches, N: tokens, C: channels, D: latent_channels
def LiSA(input, Wa, Wb, B): 
# input shape: [B,N,C], Wa shape: [N,C,D], Wb shape: [N,D], B shape: [C,D]
    qkv = linear_proj(input, channels=3C) # shape: [B,N,3C]
    query,key,value = split(qkv, [C,C,C], dim=-1)
    # query,key,value shape: [B,N,C]
    query = L2norm(query) # shape: [B,N,C]
    key = L2norm(key) # shape: [B,N,C]
    
    K_fft = rfft(key, dim=1) # shape: [B,N//2+1,C]
    Wa_fft = rfft(Wa, dim=0) # shape: [N//2+1,C,D]
    K_Wa = einsum(K_fft, Wa_fft, 'BMK,MKD->BMKD') # shape: [B,N//2+1,C,D]
    K_Wa = irfft(K_Wa, dim=1) # shape: [B,N,C,D]
    
    V_fft = rfft(value, dim=1) # shape: [B,N//2+1,C]
    Wb_fft = rfft(Wb, dim=0) # shape: [N//2+1,D]
    V_Wb = einsum(V_fft, Wb_fft, 'BMV,MD->BMVD') # shape: [B,N//2+1,C,D]
    V_Wb = irfft(V_Wb, dim=1) # shape: [B,N,C,D]
    V_Wb_B = V_Wb + B # shape: [B,N,C,D]
    
    out = einsum(query,K_Wa,V_Wb_B,'BNK,BNKD,BNVD->BNV') # shape: [B,N,C]
    out = layer_norm(out)
    out = linear_proj2(out, channels=C) # shape: [B,N,C]
    return out    
\end{lstlisting}
    \caption{
    \textbf{Pseudo-code for LiSA.} We describe the final form of LiSA described in Sec.~4.3 of the main paper.
    }
    \label{fig:LiSA_pseudo}
\end{figure}

\begin{table*}[t]
  \centering
  \renewcommand{\arraystretch}{1.5}
  % \vspace{5pt}
    %   \begin{tabu}to\textwidth{c|c|ccc}\toprule
    \scalebox{0.64}
    {
    % \begin{tabu}to\textwidth{c|c*{3}{|X[c]}}
    \begin{tabu}to\textwidth{c|c|c|c|c|c|c}
    \toprule
          & Output Size & LiSANet-I & LiSANet-S & LiSANet-B & HyLiSANet-S & HyLiSANet-B\\\midrule
    \multirow{2}[0]{*}{Stage1} & \multirow{2}[0]{*}{$\dfrac{H}{4}\times \dfrac{W}{4}$} & \multirow{2}[0]{*}{-}  & Overlap Patch Embed↓4 & Overlap Patch Embed↓4 & Overlap Patch Embed↓4 & Overlap Patch Embed↓4 \\
          &       &  & LiSA Block (64) $\times$ 2  & LiSA Block (96) $\times$ 4 & Conv Block (64) $\times$ 3 & Conv Block (96) $\times$ 3\\\midrule
    \multirow{2}[0]{*}{Stage2} & \multirow{2}[0]{*}{$\dfrac{H}{8}\times \dfrac{W}{8}$} & \multirow{2}[0]{*}{-} & Overlap Patch Embed↓2 & Overlap Patch Embed↓2 & Overlap Patch Embed↓2 & Overlap Patch Embed↓2\\
          &       &  & LiSA Block (128) $\times$ 4 & LiSA Block (192) $\times$ 8 & Conv Block (128) $\times$ 6 & Conv Block (192) $\times$ 12 \\\midrule
    \multirow{2}[0]{*}{Stage3} & \multirow{2}[0]{*}{$\dfrac{H}{16}\times \dfrac{W}{16}$} & Patch Embed↓16 & Overlap Patch Embed↓2 & Overlap Patch Embed↓2  & Overlap Patch Embed↓2  & Overlap Patch Embed↓2\\
          &       & LiSA Block (192) $\times$ 12 & LiSA Block (320) $\times$ 12 & LiSA Block (384) $\times$ 18 & LiSA Block (320) $\times$ 12 & LiSA Block (384) $\times$ 18\\\midrule
    \multirow{2}[0]{*}{Stage4} & \multirow{2}[0]{*}{$\dfrac{H}{32}\times \dfrac{W}{32}$} & \multirow{2}[0]{*}{-} & Overlap Patch Embed↓2  & Overlap Patch Embed↓2 & Overlap Patch Embed↓2 & Overlap Patch Embed↓2\\
          &       &  & LiSA Block (384) $\times$ 4 & LiSA Block (576) $\times$ 3 & LiSA Block (384) $\times$ 4 & LiSA Block (576) $\times$ 3\\\midrule
    Classifier & &\multicolumn{5}{c}{Global Average Pooling, Linear} \\\bottomrule
    \end{tabu}%
    }
      \caption{\textbf{Details of LiSANet variants.} Patch Embed↓$n$ denotes a patch embedding layer that downsamples features with a stride $n$.}\label{table:arch2} 
\end{table*}%

% \section{Additional ablation studies.}
% aaa

% \section{Visualization results.}

% t2t vit github, fig 2
\section{Additional Analyses}

\textbf{Latency.}
We demonstrate both inference and training latency in Table~\ref{tab:eff_sup}. As shown in the table, the training latency follows the tendency of the inference ones. 
LiSA becomes much faster than self-attention as the number of tokens increases due to its log-linear complexity.
We observed that training and inference latency values vary according to the FLOP size except for $7\times7$, where LiSA is slightly slower due to a higher number of sequential operations.
We employ FlashConv~\citep{fu2022hungry,fu2023flashfftconv} for FFT acceleration and adapt it to our implementation.
The kernel fusion in FlashConv addresses the I/O bottleneck by fusing the entire calculation into a single kernel and computing it in GPU SRAM.

\begin{table*}[h]
    \centering
    % \vspace{-3mm}
        % \captionsetup{width=2\columnwidth}
            \centering
            % \begin{tabular}[t]{L{2.2cm}|C{1.1cm}|C{1.1cm}|C{1.1cm}} 
 \setlength\tabcolsep{2.8pt}
            \scalebox{0.71}{
            % \begin{tabular}[t]{lcccccccc} 
    \begin{tabular}[h]{lccccccccccccccc} 
            \toprule
            block &  \multicolumn{3}{c}{$(H,W)=(7,7)$} & \multicolumn{3}{c}{$(H,W)=(14,14)$} & \multicolumn{3}{c}{$(H,W)=(28,28)$} & \multicolumn{3}{c}{$(H,W)=(56,56)$}
            & \multicolumn{3}{c}{$(H,W)=(84,84)$}\\
            % \midrule
             & FLOPs  & infer & train & FLOPs  & infer & train & FLOPs  & infer & train & FLOPs  & infer & train
             & FLOPs  & infer & train \\
             & (M)$\downarrow$ &  (ms)$\downarrow$ & (ms)$\downarrow$ & (M)$\downarrow$ &  (ms)$\downarrow$ & (ms)$\downarrow$ & (M)$\downarrow$ &  (ms)$\downarrow$ & (ms)$\downarrow$ & 
             (G)$\downarrow$ &  (ms)$\downarrow$ & (ms)$\downarrow$ & 
             (G)$\downarrow$ &  (ms)$\downarrow$ & (ms)$\downarrow$ \\
            % \midrule
            \midrule
            Self-attn~\citep{vaswani2017attention} 
            & 22.7 & \textbf{0.9} & \textbf{2.6}
            & 102.0 & 1.6 & 4.9
            & 584.0 & 13.0 & 39.6
            & 5.2 & 174.7& 531.8
            & 22.2 & OOM & OOM \\
            Convolutional attn~\citep{li2022mvitv2}
            & 23.1  & 1.7 & 4.6
            & 104.0 & 2.7 & 11.0
            & 413.0  & 8.5 & 32.1
            & 1.6  & 30.5 & 120.8
            & 3.7  & 68.0 & 266.5   \\
            \midrule
                 \rowcolor{Gray}
            LiSA (ours)
            & \textbf{21.8}  & 1.2 & 3.2
            & \textbf{87.3}  & \textbf{1.3} & \textbf{4.3}
            & \textbf{349.0} & \textbf{4.6} & \textbf{13.6}
            & \textbf{1.4}  & \textbf{19.8} & \textbf{59.2}
            & \textbf{3.1}  & \textbf{45.0} & \textbf{139.5} \\   
            \midrule
            \end{tabular}
            }
            \caption{\textbf{Comparisons among LiSA \& attention operators in FLOPs, inference latency, and training latency.} The latency is measured by an RTX A6000 (batch=32, channels=192). OOM is an abbreviation of out-of-memory.
            % FLOPs (G), the number of parameters (M), box mAP (AP$^b$) and mask mAP (AP$^m$) are shown. Note that FLOPs are measured at resolution $800\times 1280$.
            }
            \label{tab:eff_sup}
% \vspace{-5mm}
\end{table*}

\textbf{Fine-tuning on higher resolutions.}
We have verified that fine-tuning LiSANet on higher resolutions can boost image recognition accuracy. Tab.~\ref{tab:ft} summarizes the results of LiSANet-I on ImageNet. LiSANet obtains a $2.5\%$ gain when we use $384 \times 384$ resolution. While MLP-mixer models are hard to adapt to higher resolutions since they process a fixed number of tokens, LiSANet can be easily interpolated to higher resolutions due to the property of Discrete Fourier transform, where each element of the time (\ie~spatial) domain is a sampling of a continuous spectrum in the frequency domain. Since the circular embeddings $\mW^a$, $\mW^b$ can be considered as samplings of continuous spectrums, changing the resolution is equal to changing the sampling interval of spectrums~\citep{rao2021global}. Thus, LiSA can be adapted to higher resolutions by simple interpolation.
\begin{table}[t]
    \centering
        \captionsetup{width=\columnwidth}
            \centering
            % \begin{tabular}[t]{L{2.2cm}|C{1.1cm}|C{1.1cm}|C{1.1cm}} 
            \scalebox{0.85}{
            \begin{tabular}[t]{clcccc}
            % {C{0.8cm}C{2.3cm}C{1.5cm}C{1.2cm}C{1.2cm}C{0.75cm}}             
            \toprule
            index &model & Image size & FLOPs & \#params & top-1 \\
            \midrule
            \midrule
            1&LiSANet-I & $224 \times 224$ & 1.21 G & 6.36 M & 74.9 \\
            2&LiSANet-I & $384 \times 384$ & 3.62 G & 7.82 M & 77.4 \\ 
            \bottomrule
            \end{tabular}
            }
            \caption{
            \textbf{Fine-tuning to higher resolutions on ImageNet}. Image size, Top-1, accuracy (\%), FLOPs (G) and the number of paramaters (M) are shown.} \label{tab:ft} 
            % \caption{\textbf{Impact of global interactions}.} 
            % \label{table_kernel_size}
            % \afterTable
\end{table}

\begin{table}[t]
    \centering
        \captionsetup{width=0.98\columnwidth}
            \centering
            % \begin{tabular}[t]{L{2.2cm}|C{1.1cm}|C{1.1cm}|C{1.1cm}} 
            \scalebox{0.85}{
            \begin{tabular}[t]{lcccc} 
    % \begin{tabular}[t]{L{4.0cm}C{1.0cm}C{1.2cm}C{0.75cm}C{0.75cm}} 
            \toprule
            operator &  FLOPs & \#params & top-1 & top-5 \\
            \midrule
            \midrule
            Self-attention~\citep{vaswani2017attention}     & 7.36 G & 5.82 M & 18.0 & 40.9 \\
            Self-attention w/ RPE~\citep{raffel2019exploring}  & 7.36 G & 5.87 M & 24.0 & 50.0 \\
            % \midrule
            Depthwise conv ($3\times 7\times7$)~\citep{howard2017mobilenets}  & 3.75 G & 4.82 M & 33.0 & 60.9 \\
            GF layer~\citep{rao2021global}      & 3.53 G & 6.54 M & 28.7 & 40.0 \\
            Lambda convolution~\citep{bello2021lambdanetworks}   & 26.87 G & 6.34 M & 34.5 & 63.1 \\
            RSA~\citep{kim2021relational} & 72.58 G & 23.45 M & 34.1 & 62.7 \\
            % SELFYNet-TSM-R50 (ours)       & 16     &   & 40.7  \\ 
            \midrule
                 \rowcolor{Gray}
            LiSA (ours)  & 5.16 G & 8.37 M & \textbf{38.1} & \textbf{67.1} \\            
            \midrule
            \end{tabular}
            }
            % \vspace{-0.3cm}
            \caption{\textbf{Comparison with other basic operators on SS-V2}. Top-1, top-5 accuracy (\%), FLOPs (G) and the number of parameters (M) are shown.}
            \label{table_operator_video}
     % \afterTable
    % \vspace{-0.4cm}
\end{table}

\textbf{LiSA analysis on videos (Something-V2).}
% We perform experiments on the video domain to show the wide applicability of LiSA.
In Tab.~\ref{table_operator_video}, we compare different types of operators to check the feasibility of LiSA on videos, as done in Tab.~\ref{table_ablation}a for images.
The isotropic model (LiSANet-I) is trained for 60 epochs from scratch, and we adopt the uniform sampling strategy~\citep{wang2016temporal} for training and a single crop inference for testing.
We sample 8 frames per video, and the rest of the details are the same as in Sec.~\ref{sec:ablation_studies}.
% Other details are in the supplementary material.
Since structural patterns of videos, \ie, motion patterns, are important cues for recognizing video actions, the operators that learn convolutional inductive biases~\citep{zhao2020exploring,howard2017mobilenets,rao2021global,bello2021lambdanetworks} or geometric structures~\citep{kim2021relational} are more effective than their self-attention counterparts~\citep{vaswani2017attention,raffel2019exploring}.
LiSA shows even better performance on video than image, in terms of both accuracy and complexity.
While FLOPs of the attention~\citep{vaswani2017attention,zhao2020exploring} and highly-expressive~\citep{bello2021lambdanetworks,kim2021relational} operators significantly grow due to the increased number of tokens ($T\times H\times W$), LiSA remains efficient due to its log-linear complexity.

\section{Visualization}
\begin{figure*}
\centering
\begin{subfigure}{\textwidth}
    \includegraphics[width=0.98\textwidth]{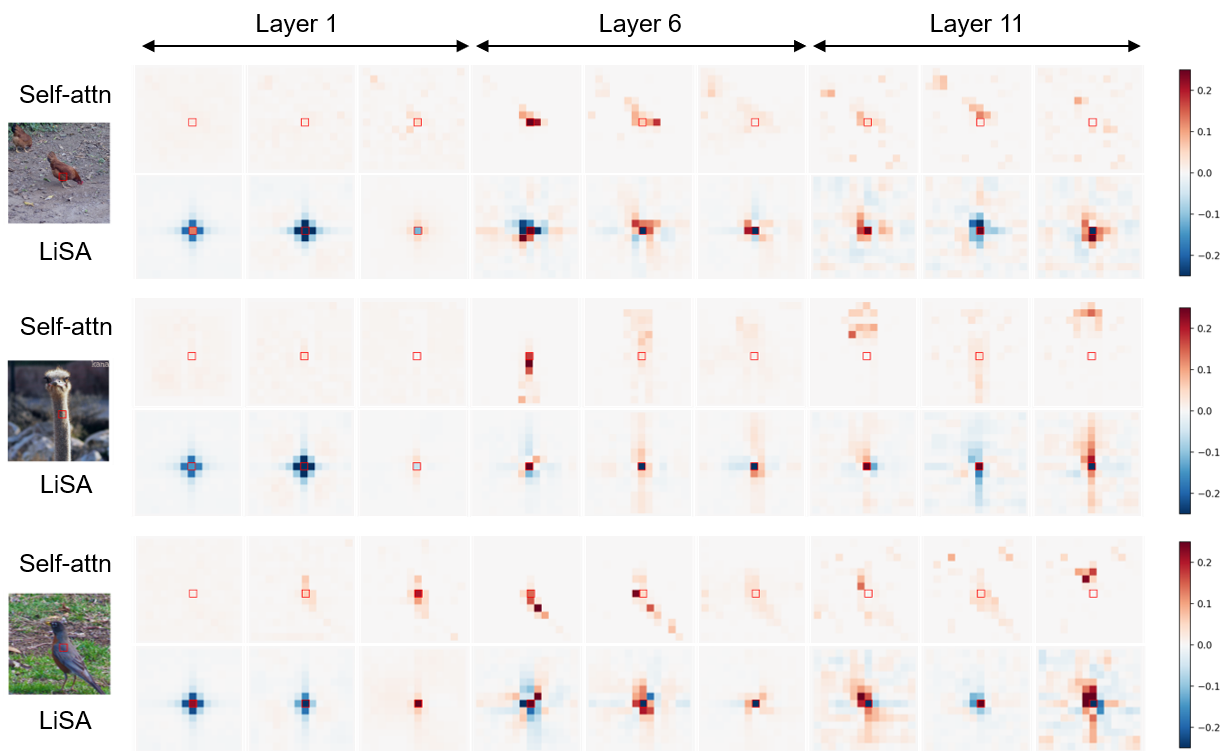}
    \caption{}
    \label{fig:viz_supp1}
\end{subfigure}
\quad
% \hfill
% \vspace{0.5cm}
\begin{subfigure}{\textwidth}
    \includegraphics[width=0.98\textwidth]{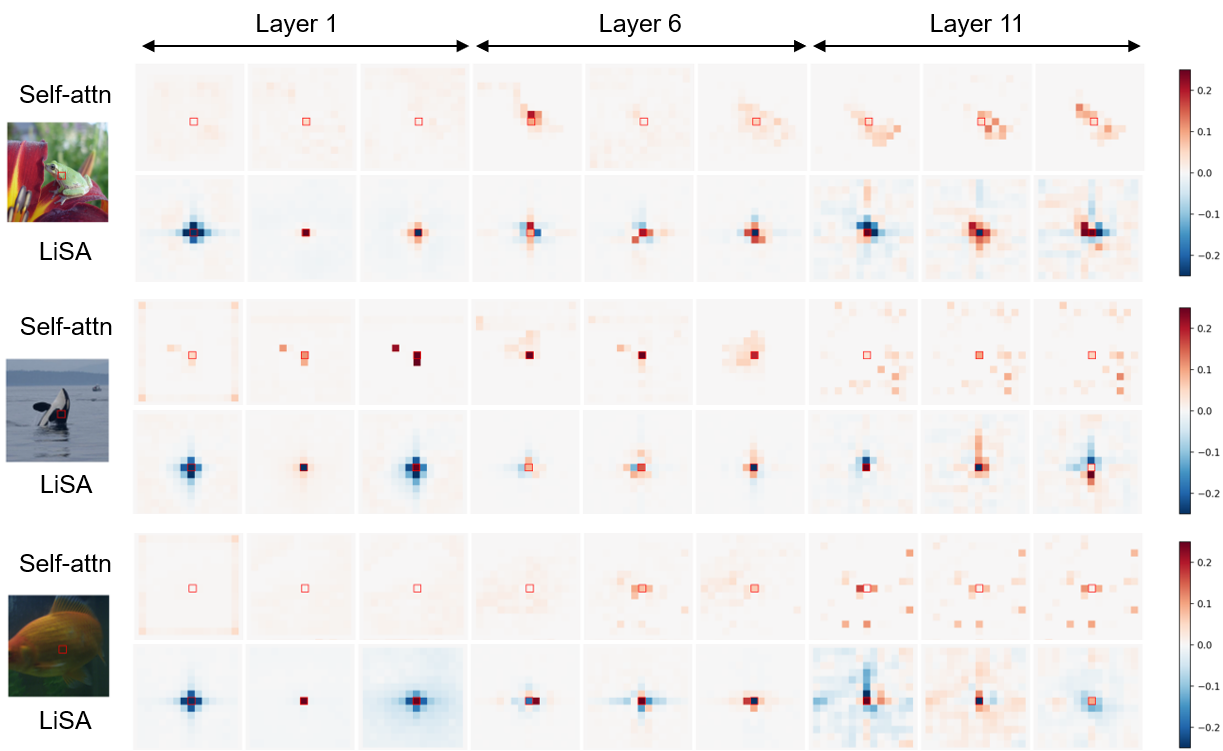}
    \caption{}
    \label{fig:viz_supp2}
\end{subfigure}
% \hfill
% \begin{subfigure}{0.4\textwidth}
%     \includegraphics[width=\textwidth]{example-image}
%     \caption{Third subfigure.}
%     \label{fig:third}
% \end{subfigure}
\caption{\textbf{Attention kernels of self-attention \& LiSA.} Attention kernels from different layers and heads are visualized. For each sample, the top row is self-attention and the bottom is LiSA. Note that the red box in the center of each subfigure is the query pixel.}
\label{fig:viz_supp}
\end{figure*}

In Fig.~\ref{fig:viz_supp}, we additionally visualize both self-attention and LiSA kernels of different layers and heads from isotropic models.
In the early layers, while self-attention kernels often fail to capture relevant context, LiSA kernels focus on encoding local features.
In the latter layers, both self-attention kernels and LiSA kernels are concentrating on the context relevant to the target object.
Therefore, LiSA kernels are much more flexible than self-attention kernels due to their ability to capture structural patterns.
% Considering that modern hybrid models~\cite{xiao2021early,dai2021coatnet,li2022uniformer,siinception}, which replace self-attention with convolution in the early layers obtain an extra accuracy gain, the behavior of LiSA kernels in the early layers seems reasonable.

\begin{figure*}[t]
\centering
    \includegraphics[width=0.8\textwidth]{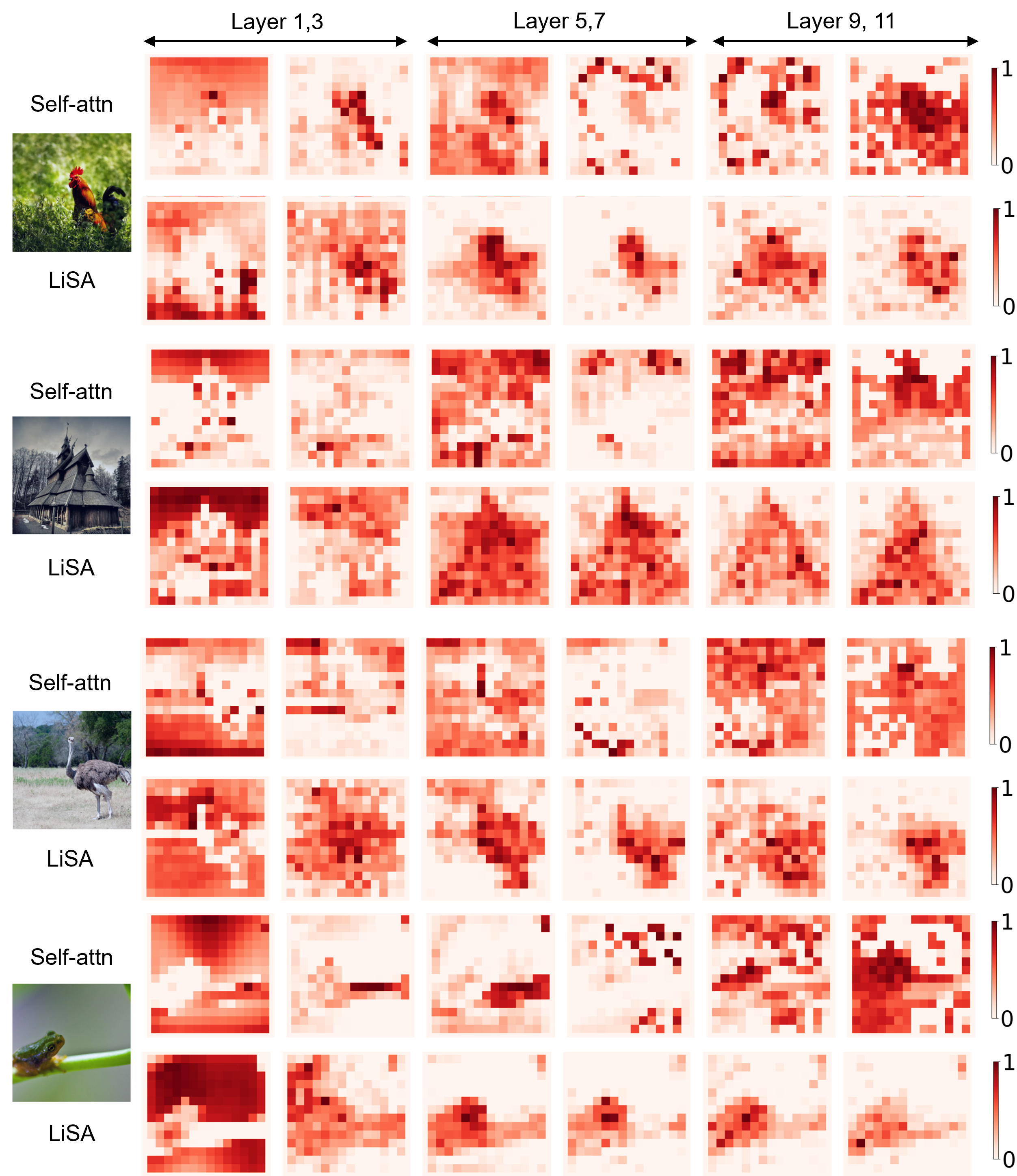}
\caption{\textbf{L2-norms of self-attention \& LiSA feature maps.} L2-norms of different intermediate feature maps are visualized. For each sample, the top row is self-attention and the bottom is LiSA.}
    \label{fig:feat_norm_supp}
\end{figure*}

In Fig.~\ref{fig:feat_norm_supp}, we visualize both self-attention and LiSA feature maps of different layers from isotropic models.
We use the L2-norm of features for each layer. 
Compared to self-attention, we observed that LiSA captures the geometric layout and positions of the main objects more effectively as the feature map progresses through the layers, verifying the effectiveness of capturing structural patterns.

\end{document}